\documentclass[10pt,journal,compsoc]{IEEEtran}

\usepackage{cite}
\ifCLASSINFOpdf

\else

\fi

\usepackage{times}  
\usepackage{helvet} 
\usepackage{courier}  
\usepackage[hyphens]{url}  
\usepackage{graphicx} 
\usepackage{graphicx}  
\usepackage{times}  
\usepackage{helvet} 
\usepackage{courier}  
\usepackage[hyphens]{url}  
\usepackage{graphicx} 
\usepackage{graphicx}  
\usepackage{amsmath,graphicx}

\usepackage{amsmath}
\usepackage{xcolor,soul,framed} 
\usepackage{xspace}
\usepackage{array}
\usepackage{eqparbox}
\usepackage{url}
\usepackage{longtable}
\usepackage{lipsum}
\usepackage{blindtext}
\usepackage{makecell}
\usepackage{mathtools}
\usepackage{commath}
\usepackage{multirow}
\usepackage{tabularx}
\usepackage[normalem]{ulem}
\usepackage{xspace}
\usepackage{algorithm}
\usepackage{algpseudocode}
\usepackage{amsfonts}
\usepackage{placeins}

\usepackage[normalem]{ulem}

\usepackage{array}
\usepackage{color, colortbl}
\definecolor{Gray}{gray}{0.9}
\usepackage{eqparbox}
\usepackage{url}
\usepackage{lipsum}
\usepackage{blindtext}
\usepackage{makecell}

\usepackage{mathtools}
\usepackage{commath}
\usepackage{multirow}
\usepackage{tabularx}
\usepackage[normalem]{ulem}
\usepackage{xspace}
\usepackage{algorithm}
\usepackage{algpseudocode}
\usepackage{amsfonts}
\usepackage{placeins}
\usepackage{booktabs}
\usepackage[normalem]{ulem}
\usepackage{pifont}
\newcommand{\xmark}{\ding{55}}%

\newcolumntype{C}{>{\hsize=\dimexpr0.5\hsize+8\tabcolsep+\arrayrulewidth\centering\relax}X}

\DeclareMathOperator*{\argmax}{arg\,max}
\DeclareMathOperator{\cref}{ref}
\DeclareMathOperator{\emb}{emb}
\DeclareMathOperator{\pos}{pos}

\DeclareMathOperator{\queue}{queue}
\DeclareMathOperator{\bmm}{bmm}
\DeclareMathOperator{\dist}{dist}

\DeclareMathOperator{\proto}{proto}
\DeclareMathOperator{\cat}{cat}

\newcommand{\convMatrix}[3]{
\begin{math}
\begin{bmatrix}
\text{Conv1D},\text{kernel}(#1), \text{output channels}(#2) \\
\text{BatchNorm1D},\text{output channels}(#3) \\ 
\text{LeakyReLU},\text{slope(0.3)}\ 
\end{bmatrix} 
\end{math}
}

\newcommand{\convMatrixc}[1]{
\begin{math}
\begin{bmatrix}
\text{Linear}, 64 \\
\text{ReLU}, - \\ 
\text{Dropout},0.5\\ 
\text{Linear},#1\ 
\end{bmatrix} 
\end{math}
}

\usepackage{listings}
\usepackage[frozencache=true,cachedir=minted-cache]{minted} 

\usepackage[pagebackref=true,breaklinks=true,colorlinks,bookmarks=false]{hyperref}
\hypersetup{
    colorlinks=true,
    filecolor=magenta,      
    urlcolor=cyan,
}

\begin{document}
\title{Partial Label Learning for Emotion Recognition from EEG}

\author{Guangyi Zhang and Ali Etemad \IEEEmembership{Senior Member, IEEE} \thanks{Guangyi Zhang and Ali Etemad are with the Department of Electrical and Computer Engineering, Queen's University, Kingston, ON, Canada K7L3N6. (e-mail: \{guangyi.zhang, ali.etemad\}@queensu.ca)}}

\maketitle

\begin{abstract}
Fully supervised learning has recently achieved promising performance in various electroencephalography (EEG) learning tasks by training on large datasets with ground truth labels. However, labeling EEG data for affective experiments is challenging, as it can be difficult for participants to accurately distinguish between similar emotions, resulting in ambiguous labeling (reporting multiple emotions for one EEG instance). This notion could cause model performance degradation, as the ground truth is hidden within multiple candidate labels. To address this issue, Partial Label Learning (PLL) has been proposed to identify the ground truth from candidate labels during the training phase, and has shown good performance in the computer vision domain. However, PLL methods have not yet been adopted for EEG representation learning or implemented for emotion recognition tasks. \textcolor{black}{In this paper, we adapt and re-implement six state-of-the-art PLL approaches for emotion recognition from EEG on two large emotion datasets (SEED-IV and SEED-V). These datasets contain four and five categories of emotions, respectively.} \textcolor{black}{We evaluate the performance of all methods in classical, circumplex-based and real-world experiments.} The results show that PLL methods can achieve strong results in affective computing from EEG and achieve comparable performance to fully supervised learning. We also investigate the effect of label disambiguation, a key step in many PLL methods. The results show that in most cases, label disambiguation would benefit the model when the candidate labels are generated based on their similarities to the ground truth rather than obeying a uniform distribution. \textcolor{black}{This finding suggests the potential of using label disambiguation-based PLL methods for circumplex-based and real-world affective tasks.}
\end{abstract}

\section{Introduction} \label{sec:intro}
Emotion has been widely recognized as a mental state and a psycho-physiological process that can be associated with various brain or physical activities \cite{dalgleish2004emotional,koelstra2011deap,zhang2021deep}. It can be expressed through various modalities, such as facial expression, hand gestures, body movement, and voice, with various levels of intensity \cite{peelen2010supramodal}. Emotion plays an important role in our daily life by affecting decision-making and interactions \cite{turner2006sociological}; it is thus vital to enable computers to identify, understand, and respond to human emotions \cite{picard2000affective} for better human-computer interaction and user experience. In recent decades, among non-invasive physiological measurements such as electrocardiography \cite{sarkar2020self}, electromyography \cite{jerritta2011physiological}, and electrodermal activity \cite{bhatti2021attentive}, Electroencephalography (EEG) has been widely used for human emotion recognition due to its direct reflection of brain activities \cite{koelstra2011deap,zhang2020rfnet,zhang2021distilling,liu2021comparing} 

\textcolor{black}{One of the critical challenges in EEG labeling is the lack of reliability in self-assessment of emotions \cite{correa2018amigos}. For example, while participants may easily distinguish between \textit{dissimilar} emotions or feelings such as `happy' vs. `fear', they often struggle to distinguish similar ones, for instance, `disgust' vs. `fear'. \textcolor{black}{This difficulty in accurately identifying emotions may result in unreliable self-assessment reports \cite{correa2018amigos}.} Since accurate labeling of emotions is difficult and expensive, participants should ideally be allowed to report multiple possible emotions for a single EEG instance. However, fully supervised learning models, which rely on precise, singular labels, are not designed to handle this ambiguity \cite{xia2023towards}.} 

\textcolor{black}{Recently, \textit{self- and semi-supervised learning} approaches have been explored for emotion recognition using EEG to address the scarcity of labeled EEG data \cite{zhang2021deep,zhang2022holistic,zhang2022parse,wang2023self,liu2024self,fu2024gmaeeg, jiang2025neurolm, wang2025cbramod}. \textcolor{black}{While such approaches can improve model performance when faced with limited labels, they do not address the challenge of `ambiguous' labeling. Self-supervised and semi-supervised learning assume that existing labels are accurate and reliable, which is not always the case in emotion self-assessment, particularly when emotions are similar or closely positioned on the arousal and valence dimensions, as described by Russell's circumplex model \cite{russell1980circumplex}.}}

\begin{figure}
    \begin{center}
    \includegraphics[width=0.48\textwidth]{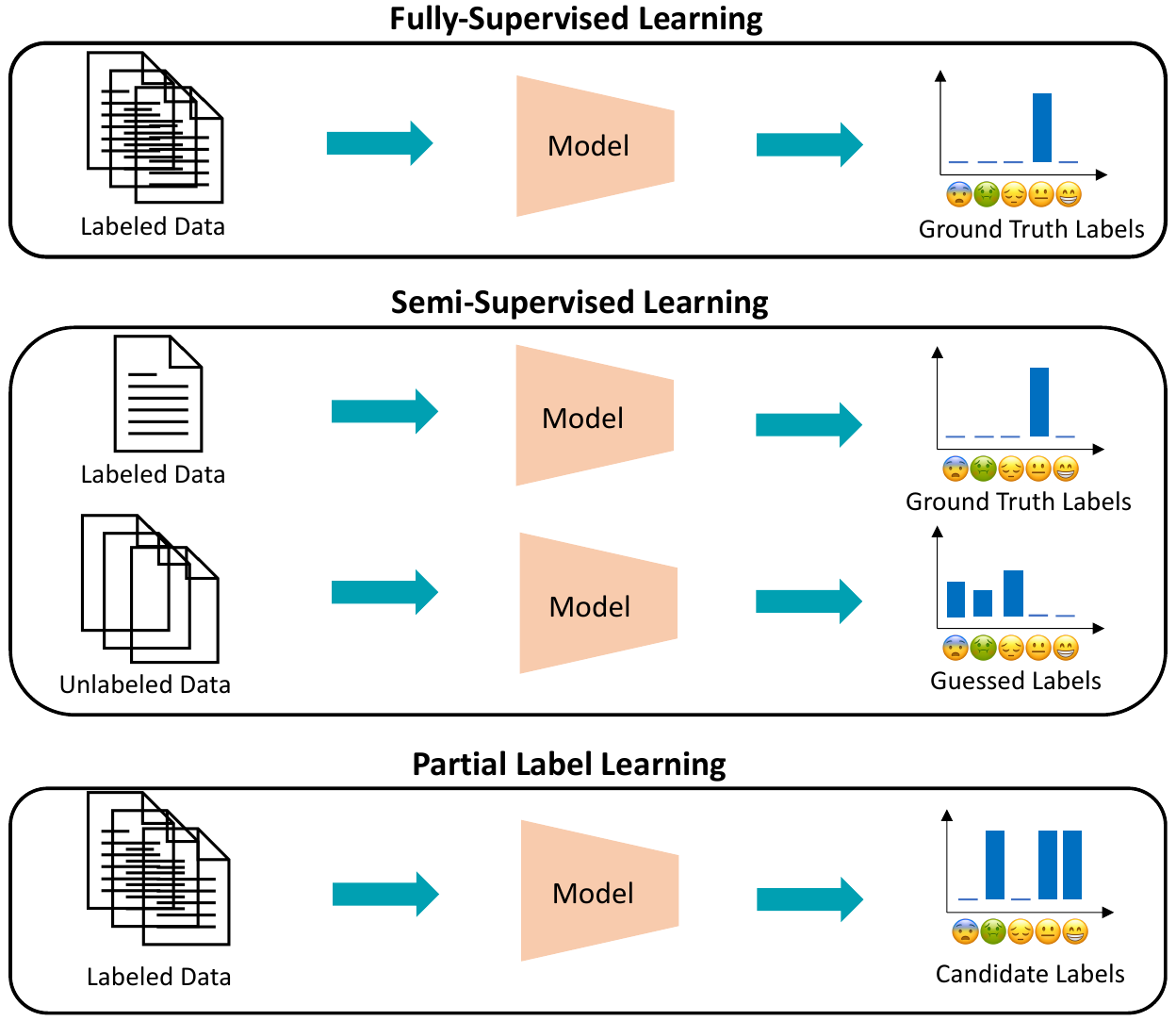} 
    \caption{A description of full-supervised learning, semi-supervised learning and partial label learning for emotion recognition is presented.}
    \label{sl_comparison}
    \end{center}
\end{figure}

To tackle the above-stated problems, we propose a framework to allow participants to report multiple possible emotions if they are uncertain about their affective state during the self-assessment stage. \textcolor{black}{This notion, referred to as \textit{\textbf{Partial Label Learning (PLL)}}, has shown effectiveness in handling label ambiguity in domains such as computer vision, yet it remains unexplored in EEG-based emotion recognition. In contrast to self-supervised and semi-supervised learning methods that typically rely on reliable and accurate labels, the proposed PLL approach is specifically designed to tackle label ambiguity by incorporating multiple candidate labels and applying a label disambiguation mechanism during training.} However, ambiguous labeling often causes performance degradation in deep learning algorithms as the ground truth is hidden within the candidate labels during the training phase \cite{wang2022pico}, making PLL a challenging area. \textcolor{black}{
In this paper, we explore several state-of-the-art PLL techniques \cite{seo2021power,lv2020progressive,zhang2022exploiting,wen2021leveraged,wu2022revisiting,wang2022pico} originally proposed in the field of computer vision. For the first time, we adapt these methods for EEG-based emotion recognition, demonstrating their potential in a novel application area. We also adapt two large-scale emotion EEG datasets, SEED-IV \cite{zheng2018emotionmeter} and SEED-V \cite{liu2021comparing} by substituting the original labels (ground truth) with `candidate' label \textit{sets}, allowing the evaluation of PLL techniques for EEG-based emotion recognition.} 
This work comprehensively compares and analyzes these techniques to understand the viability of using PLL for EEG-based emotion recognition. An overview of the core concept behind PLL is provided in Figure \ref{sl_comparison}.

Our contributions in this paper are as follows. \textcolor{black}{(\textbf{1}) For the first time, we introduce PLL to address the challenge of ambiguous EEG labeling in emotion recognition tasks, offering a novel approach to manage uncertainty in self-assessed emotions. (\textbf{2}) We re-implement and adapt six recent state-of-the-art PLL algorithms on two large EEG datasets, providing a comprehensive comparison of these methods for emotion recognition. (\textbf{3}) We explore different strategies for generating candidate labels, including simulation of circumplex-based and real-world emotional ambiguity. By introducing various levels of label uncertainty, we systematically evaluate the impact of these strategies on the performance of the six PLL methods, providing insights into their robustness and practical effectiveness in learning from EEG data with ambiguous emotional labels.}

The rest of this paper is organized as follows. In the next section, a literature survey on EEG learning for affective computing is provided followed by a summary of PLL methods proposed in the literature. In Section \ref{sec:method}, the problem statement and the details of the PLL techniques adopted for EEG are provided . The datasets used in this work are then presented in Section \ref{sec:experiment}, along with implementation details and experiment setup. In Section \ref{sec:result}, the detailed results and analysis are provided. And finally Section \ref{sec:conclusion} presents the concluding remarks.

\begin{figure*}[!ht]
    \begin{center}
    \includegraphics[width=0.8\textwidth]{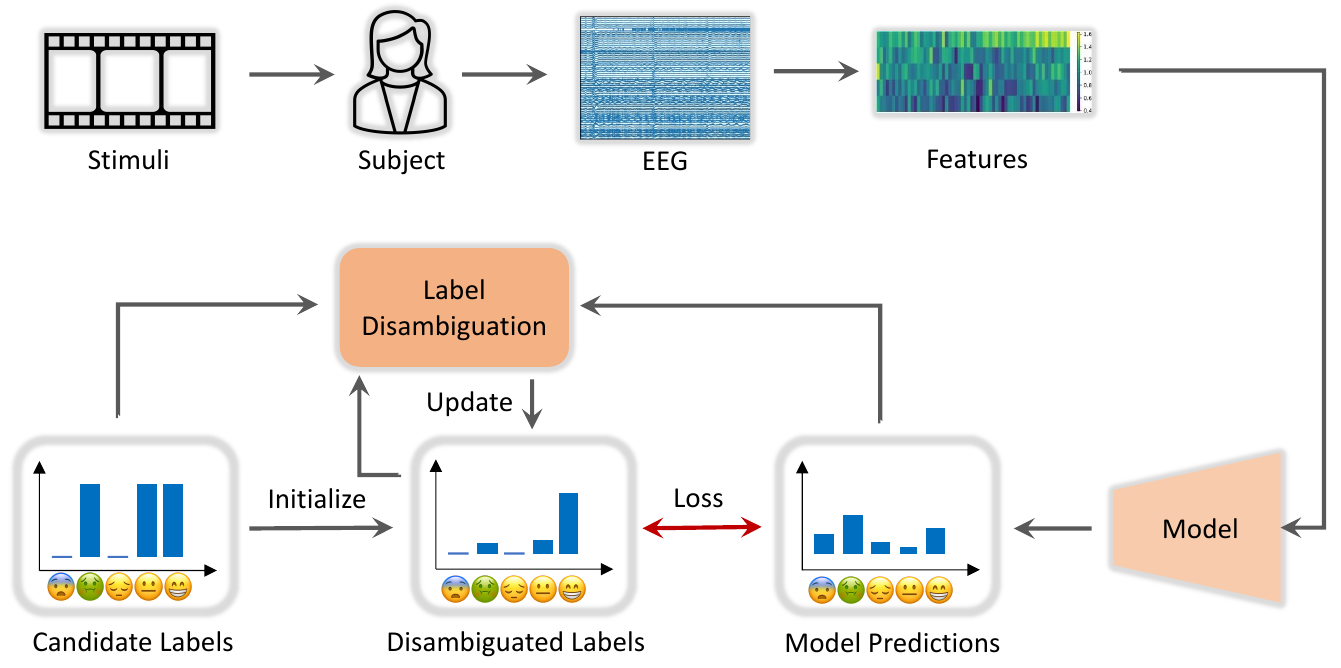} 
    \caption{A general EEG-based PLL framework for emotion recognition is presented.}
    \label{overview}
    \end{center}
\end{figure*}

\section{Background} \label{sec:background}
\textcolor{black}{In this section, a literature review of works in the area of deep EEG learning is first provided, followed by a review of recent works in the area of PLL, which have been originally proposed for computer vision applications.}

\subsection{Deep EEG Representation Learning}
\textcolor{black}{Deep learning techniques, such as deep belief networks \cite{zheng2015investigating}, fully connected neural networks \cite{zhang2020rfnet}, Convolutional Neural Networks (CNNs) \cite{lawhern2018eegnet}, capsule networks \cite{zhang2021capsule,zhang2021distilling}, Recurrent Neural Networks (RNNs) \cite{roy2019chrononet}, long short-term memory networks \cite{zhang2019classification}, and graph neural networks \cite{song2018eeg,zhong2020eeg,li2022residual} 
have been widely used for EEG-based fully-supervised tasks, such as motor imagery or movement classification and emotion recognition \cite{craik2019deep,roy2019deep}.} These deep learning techniques outperformed classical statistical algorithms and conventional machine learning methods in most tasks, as they are able to learn non-linear and more complex patterns and focus on task-relevant features \cite{zhang2019classification,zhang2020rfnet,zhang2021capsule}. CNNs are the most popular deep learning backbones for EEG learning \cite{craik2019deep,roy2019deep,lawhern2018eegnet}. \textcolor{black}{Lately, transformers with various pre-training strategies have achieved state-of-the-art results in emotion recognition tasks across multiple EEG datasets \cite{yi2024learning,jianglarge,jiang2024neurolm}.}

Deep EEG representation learning frameworks have also shown good performance in semi-supervised tasks, where only a few EEG annotations are available during training \cite{zhang2021deep,zhang2022holistic}. An attention-based recurrent autoencoder was proposed for semi-supervised EEG learning \cite{zhang2021deep}. Furthermore, 
state-of-the-art semi-supervised techniques originally developed for computer vision tasks, such as MixMatch \cite{berthelot2019mixmatch}, FixMatch \cite{berthelot2019mixmatch} and AdaMatch \cite{berthelot2021adamatch}, have been adapted for EEG learning and obtained promising results in emotion recognition tasks. A novel pairwise EEG representation alignment method (PARSE), based on a lightweight CNN backbone, was lately proposed and achieved state-of-the-art results in semi-supervised tasks on multiple publicly available large-scale affective datasets \cite{zhang2022parse}. More importantly, PARSE achieved similar performance to fully-supervised models trained on large-scale labeled datasets with substantially fewer labeled samples, demonstrating the superiority of deep semi-supervised EEG learning in the face of a scarcity of labeled data \cite{zhang2022parse}. \textcolor{black}{More recently, to address the issue of label scarcity, DS-AGC employed multi-domain adaptation and graph contrastive learning to extract EEG representations in a semi-supervised manner, achieving strong results in cross-subject emotion recognition experiments \cite{ye2024semi}.}

\subsection{Partial Label Learning}

PLL algorithms have been lately used to tackle the challenges of label ambiguity and achieved promising performance in a variety of image classification tasks with ambiguous labels. For instance, in \cite{seo2021power}, a deep naive partial label learning model was proposed based on the assumption that the distribution of candidate labels should be uniform since ground truth is unknown \cite{jin2002learning}. In addition to this naive approach, a number of other frameworks have been lately developed to rely on a process called `label disambiguation', which refines the distribution of candidate labels by updating it in each training iteration. This process of combining label disambiguation and model classification works as an Expectation-Maximization (EM) algorithm as shown in \cite{jin2002learning}. Specifically, the candidate labels' distribution is initially assumed to be uniform. In the first iteration, a model is trained on the uniformly distributed candidate labels. Then, the candidate labels are disambiguated based on the trained model's predictions. In the next iteration, the model is trained on the disambiguated candidate labels, and the process repeats during the training phase \cite{jin2002learning,lv2020progressive}.

In \cite{zhang2022exploiting}, a method was proposed for label disambiguation that utilized the importance of each output class, rather than relying solely on the model's predictions. To mitigate the possible negative effect of training on false positive labels, in \cite{wen2021leveraged,wu2022revisiting}, a method was proposed to leverage both candidate and \textit{non-candidate} labels with the label disambiguation process. Furthermore, in \cite{wang2022pico}, a prototype-based label disambiguation method was proposed, which was used in combination with supervised contrastive learning \cite{khosla2020supervised}, a technique that relies on contrastive loss to train a Siamese-style network. Specifically, prototype-based label disambiguation is first used to guess the ground truth and generate accurate positive pairs for contrastive learning. Then, the contrastively learned embeddings, in turn, could better guide the label disambiguation process. Thus, these two components are mutually beneficial during the iterative training, leading the PLL framework to achieve state-of-the-art results in multiple vision tasks \cite{wang2022pico}.

\section{PLL Methods} \label{sec:method}
\textcolor{black}{In this section, we first provide a clear description of the problem setup, followed by the details of the PLL methods used for EEG-based emotion recognition in this study.} \textcolor{black}{While we build upon existing PLL techniques, we introduce several key modifications and adaptations to enhance performance specifically in the EEG domain.}

\subsection{Problem Setup} \label{sec: problem setup}
Let's denote $\mathcal{X}$ and $\mathcal{Y} \in \{1,2,...k\}$ as the input EEG feature space and the emotion label space, respectively, where $k$ is the total number of emotion categories. Accordingly, $x \in \mathcal{X}$ and $y \in \mathcal{Y}$ represent EEG samples and emotion ground truth labels. 
For a classification problem with ambiguous labels, \textcolor{black}{the collection of all subsets in $\{1,2,...k\}$, $2^{\mathcal{Y}}\setminus\emptyset\setminus\mathcal{Y}$ has been used instead of $\mathcal{Y}$ as the label pace.}
The training set $D = \{(x_i, Y_i)\}_{i=1}^N$ consists of normalized EEG input $x_i \sim[0,1]$, \textcolor{black}{and candidate label $Y_i \subset 2^{\mathcal{Y}}\setminus\emptyset\setminus\mathcal{Y}$.} The ground truth $y_i$ is concealed in the candidate labels as $y_i \in Y_i$.

In classical PLL literature, candidate labels are assumed to be generated independently, uniformly, and randomly. \textcolor{black}{Thus, for the distribution of each candidate label set, $\mathrm{P}(Y_i\mid x_i, y_i, q) = \prod_{s \neq y_i} q, \forall s \in[1,k]$, where $\mathrm{P}(Y_i\mid x_i, y_i, q)$ represents the label ambiguity and $q<1$ represents the degree of label ambiguity.} The goal of PLL is to construct a robust multi-class classifier by minimizing the divergence between model output and candidate labels. In this study, we aim to evaluate the effectiveness of various PLL algorithms for EEG representation learning when emotion labels with different levels of ambiguity are provided.

\subsection{Method Overview} \label{sec: solutions overview}
A general PLL framework for emotion recognition is illustrated in Figure \ref{overview}. In a typical EEG-based emotion recognition experiment setup, EEG recordings are collected from a participant's brain scalp while they are watching an emotion-related video clip. Then, candidate labels are generated based on 
the participant's self-assessment. These candidate labels are then used to train a model with the features extracted from EEG recordings. \textcolor{black}{In the context of partial label learning, `label disambiguation' is an additional mechanism beyond the softmax operation used by the learner or backbone itself.} This mechanism is often based on \textit{uniformly distributed} candidate labels ($\hat{Y}$) or the disambiguated labels obtained from the last training iteration ($\vec{Y}$) and model predictions ($p_m(x)$). The disambiguated labels ($\vec{Y}$) are updated in each training iteration, and the model is trained by minimizing the divergence between the model predictions and the disambiguated labels. When label disambiguation is not used, the divergence will consistently be minimized between model predictions and the candidate labels (with uniform distribution) during the entire training phase. Training is performed by batches, where $x_b$ denotes input samples of each batch.

We identify six state-of-the-art PLL techniques from the literature, which we re-implement and adapt for emotion recognition from EEG. These methods are Deep Naive Partial Label (DNPL) learning \cite{seo2021power}, PROgressive iDENtification (PRODEN) true Labels for PLL \cite{lv2020progressive}, Class Activation Value Learning (CAVL) for PLL, \cite{zhang2022exploiting}, Leveraged Weighted (LW) loss for PLL \cite{wen2021leveraged}, revisiting Consistency Regularization (CR) \cite{wu2022revisiting} for deep PLL, and Partial label learning with COntrastive label disambiguation (PiCO) \cite{wang2022pico}, which are described in the following sub-sections.

\subsection{DNPL}
In DNPL \cite{seo2021power}, since the ground truth is unknown \cite{jin2002learning}, the candidate labels are assumed to have uniform distribution as follows: 
\begin{equation}\label{eq: uniform}
\overline{\hat{Y}_s}= \frac{1}{|Y|} ~~~~\text{if}~~s \in Y, ~~~0~~  \text{otherwise}, 
\end{equation} 
where $\hat{Y}_s=1, \forall s \in Y$.
The DNPL model is simply trained by minimizing the divergence between the model's predictions and the uniformly distributed candidate labels, as:
\begin{equation}\label{eq: DNPl}
\mathcal{L}_{naive} = - \frac{1}{|x_b|}\sum_{i=1}^{|x_b|} \log(\nu(\hat{Y} \phi(p_m(x_i)))),
\end{equation} 
where $\phi(.)$ denotes the softmax operation, and $\nu$ is the clamp operator which limits the output in the range of $[0,1]$.

\subsection{PRODEN}
PRODEN was proposed to refine candidate labels through a label disambiguation process \cite{lv2020progressive}. Specifically, the process obtains disambiguated labels by multiplying model predictions with candidate labels, as: $\vec{Y} = p_m(x) \hat{Y}$. Following, the model is trained by minimizing cross-entropy between the model's predictions and the disambiguated labels, as: 
\begin{equation}\label{eq: PRODEN}
\mathcal{L}_{ce} = -\frac{1}{|x_b|}\sum_{i=1}^{|x_b|}  \vec{Y} ~\log (\phi(p_m(x_i))).
\end{equation}
Here, when the label disambiguation process is not used, $\vec{Y}$ will be replaced by $\hat{Y}$.

\subsection{CAVL}
In \cite{zhang2022exploiting}, CAVL was proposed to disambiguate candidate labels by focusing on the importance score of each label in the model's predictions. Specifically, inspired by \cite{selvaraju2017grad}, CAVL uses gradient flow ($v^s$) of the network output's log probability as a measurement of the importance of each label, which is shown as:
\begin{equation}\label{eq: CAVL}
\begin{split}
    v^s &= \bigg| \frac{\partial(-\log(\phi^s(p_m(x))))}{\partial p_m(x)} \bigg|\partial \phi^s(p_m(x)) \\
    &= \big|\phi^s(p_m(x))-1\big|\phi^s(p_m(x)), \forall s \in [1,k].
\end{split}
\end{equation}
As suggested in \cite{zhang2022exploiting}, $\phi(p_m(x))$ is replaced by $p_m(x)$ since $p_m(x)$ contains more information. Consequently, $\hat{v}^s = \big|p_m^{s}(x_i)-1\big|p_m^{s}(x_i)$ and disambiguated label $\vec{Y} = \argmax (\hat{v}^s \overline{\hat{Y}})$. The same cross-entropy loss in (Eq. \ref{eq: PRODEN}) is used for model training. When label disambiguation is not used, same as in PRODEN, $\vec{Y}$ will be replaced by $\hat{Y}$.

\subsection{LW} \label{method: LW}
Most existing PLL algorithms only focus on learning from candidate labels while ignoring non-candidate ones \cite{seo2021power,lv2020progressive,zhang2022exploiting,wang2022pico}. However, a model could be misled by the false positive labels if it only relies on the candidate label set \cite{wu2022revisiting}. To mitigate this possible negative effect, in \cite{wen2021leveraged}, both candidate and non-candidate labels have been used for model training using an LW loss. To do so, first, the sigmoid loss $\sum_{s\notin Y}\varphi(p_m(x))$ ($\varphi(.)$ representing sigmoid operation) or negative log-likelihood loss $-\sum_{s\notin Y} \log(1-\phi(p_m(x)))$ is applied on non-candidate labels, in order to discourage the model predictions to be among the non-candidate labels. Next, the modified sigmoid loss $\sum_{s\in Y} \varphi(-p_m(x))$ or negative log-likelihood loss $-\sum_{s\in Y}\log(\phi(p_m(x)))$ is employed for model training on candidate labels. Finally, the model is trained on both candidate and non-candidate labels. The total loss function using the sigmoid function and cross-entropy loss (negative log-likelihood) are shown in Eq. \ref{eq: LW-sigmoid} and Eq. \ref{eq: LW-CE}, respectively. 
\begin{equation}\label{eq: LW-sigmoid}
\mathcal{L}_{ce}^{lw} = -\frac{1}{|x_b|}\sum_{i=1}^{|x_b|}\big[ \sum_{s\in Y} w_s \varphi(-p_m(x_i)) + \beta \sum_{s\notin Y} w_s \varphi(p_m(x_i)) \big],
\end{equation} 
\begin{equation}\label{eq: LW-CE}
\begin{split}
\mathcal{L}_{sig}^{lw} & =  - \frac{1}{|x_b|}\sum_{i=1}^{|x_b|}\big[\sum_{s\in Y} w_s \log(\phi(p_m(x_i)))\\ & + \beta \sum_{s\notin Y} w_s \log(1-\phi(p_m(x_i))) \big].
\end{split}
\end{equation}
A trade-off parameter $\beta=0, 1, 2$ is applied as suggested in \cite{wen2021leveraged}. The weights $w_s$ are the normalized model predictions, as $w_s = \phi(p_m(x_i)), \forall s \in Y$ and $w_s = \phi(p_m(x_i)), \forall s \notin Y$. The weights ($w_s$) are updated chronologically and used to assign more weights to learn the candidate labels which are more likely to be the ground truth and the more confusing non-candidate labels. Note that $w_s=1$ when label disambiguation is not used. 

\subsection{CR}
In \cite{wu2022revisiting}, both candidate and non-candidate labels are leveraged in a similar manner to the LW method \cite{wen2021leveraged}, where modified negative log-likelihood loss is applied on non-candidate labels, as $\mathcal{L}_s = - \frac{1}{|x_b|} \sum_{i=1}^{|x_b|} \sum_{s \notin Y} \log(1-\phi(p_m(x_i)))$. In \cite{wu2022revisiting}, instead of using cross-entropy loss, 
a consistency regularization method has been proposed for learning candidate labels. Consistency regularization encourages a model to have consistent predictions on the original data or its perturbed versions \cite{samuli2017temporal,berthelot2019mixmatch,zhang2022parse}. As suggested in \cite{wu2022revisiting}, consistency regularization is applied by minimizing the divergence between each augmentation of an instance and its \textit{refined} label. To do so, the refined label is first obtained based on the network predictions of the original data and its different augmentations, as: 
\begin{equation}\label{eq: CR disambiguation}
\vec{Y}_{s} = \prod_{i=1}^{3} \phi(p_m(A_i(x_b)))^{1/3}, \forall s \in Y
\end{equation} 
where $A_i$ presents the original, weakly, and strongly augmented data, as $x,A_w,A_s$ when $i=1,2,3$. Additive Gaussian noise has been commonly used as an effective data augmentation method in recent EEG studies \cite{li2019channel,luo2020data,zhang2022holistic,zhang2022parse}. Therefore, the augmented data are produced as $\mathcal{A}_{s/w}(x) = x + \mathcal{N}(\mu, \sigma)$, where $\mathcal{N}$ represents a Gaussian distribution. As suggested in \cite{zhang2021capsule,zhang2022parse}, a mean value ($\mu$) of $0.5$ is chosen, along with  standard deviation values ($\sigma$) of $0.8$ and $0.2$ for strong ($\mathcal{A}_{s}$) and weak ($\mathcal{A}_{w}$) augmentations respectively. Following, the divergence between different augmentations of an instance and its refined label is minimized, 
\begin{equation}\label{eq: CR unsupervised loss}
\mathcal{L}_u = \frac{1}{|x_b|}\sum_{i=1}^{|x_b|} \sum_{j=1}^{3} D_{KL}(\vec{Y}'||\log(\phi(p_m(A_j(x_i))))),
\end{equation} where $D_{KL}$ denotes Kullback-Leibler (KL) divergence and $\vec{Y}'$ denotes the normalized disambiguated labels, as $\vec{Y}' = \vec{Y}/ \sum_{s \in Y} \vec{Y}$.
Finally, the total loss consists of a supervised loss applied on non-candidate labels and a consistency regularization term applied on candidate labels,
\begin{equation}\label{eq: CR loss}
\mathcal{L}^{cr} = \mathcal{L}_s+ \eta \mathcal{L}_u,
\end{equation} 
where $\eta= t/T$ is the warm-up function. $t$ and $T$ represent the current epoch and the total training epochs, respectively. Moreover, $\eta$ is set to 0 when label disambiguation is not used.

\subsection{PiCO} \label{method: PiCO}
PiCO was proposed to incorporate contrastive learning and a prototype-based label disambiguation method for effective partial label learning \cite{wang2022pico}. PiCO employs a similar framework with MoCo \cite{he2020momentum}, which includes a query network $p_m$ and a key network $p'_m$ sharing the same encoder architecture, where $p'_m$ uses a momentum update with $p_m$ \cite{he2020momentum}. Following, the details of contrastive learning and the prototype-based label disambiguation technique are described.

\textbf{Contrastive Learning.} Contrastive learning has achieved promising performance in both supervised and unsupervised learning tasks by maximizing the similarity between learned representations of a positive pair \cite{khosla2020supervised,chen2020simple,he2020momentum, zhou2025brainuicl}. In supervised contrastive learning, a positive pair includes samples from the same class, while a negative pair contains examples belonging to different classes. In unsupervised contrastive learning, a positive pair often consists of a sample and its augmentation, while a negative pair consists of a sample and a randomly chosen instance. 

Since the ground truth is unknown in partial label learning tasks, contrastive learning could be performed in an unsupervised manner. To do so, a positive pair is first constructed using an instance ($x_i$) and its weak augmentation ($A_w(x_i)$). Then, a pair of L$2$ normalized embeddings is obtained from the query and key networks, as $Q_i=||p_{\emb}(x_i)||_2$, $K_i=||p'_{\emb}(A_w(x_i))||_2 \in (0, 1)^{N_{\emb}}$, where $p_{\emb}$ and $p'_{\emb}$ denotes the embeddings obtained from encoders $p_m$ and $p'_m$, respectively. $N_{\emb}=64$ denotes the dimension of the embeddings. Following, the negative sample pool is initialized as $\queue \in \mathbb{R}^{[N_q \times k]}, \text{where} \queue_{i,j} \sim \mathcal{N}(0,1), \forall i \in [1, N_q], \forall j \in [1,N_{\emb}]$, as suggested in \cite{he2020momentum}.
The length of the pool, $N_q$, is set to $N_q=1000$, according to the total number of EEG training samples.
Next, the unsupervised contrastive loss is calculated as:
\begin{equation}\label{eq: unsupervised contrastive loss}
\mathcal{L}_{u}^{ct} = - \frac{1}{|x_b|} \sum_{i=1}^{|x_b|} \log \frac{\exp((\bmm(Q_i, K_i)/\tau)}{\sum_{i=1}^{|x_b|}\sum_{1}^{N_q} \exp(Q_i \queue^\top/\tau)},
\end{equation} 
where $\bmm$ denotes the batch matrix multiplication operator such that $\bmm(Q_i, K_i) \in \mathbb{R}^{[|x_b|, 1]}$. The temperature hyper-parameter $\tau$ is set to 0.07, as suggested in \cite{wang2022pico}.

Unsupervised contrastive learning remains challenging since the false negative samples in the negative samples pool may cause performance degradation \cite{khosla2020supervised,chen2022incremental}. To tackle this challenge, one solution is to construct a positive pair consisting of two samples whose \textit{guessed} labels are the same. To do so, the pseudo-label is first estimated as $\vec{Y}_i = \phi{(p_m(x_i))} \overline{\hat{Y}_i}$. 
The contrastive representation pool is then constructed as $\queue^{+}= \cat(K, \queue)$, where cat$(.)$ denotes a concatenation operation. The corresponding pseudo-label pool is denoted as ${Y^{\queue^{+}} = \cat(\overline{\hat{Y}}, Y^{\queue})}$, where $Y^{\queue}\sim \mathcal{N}(0,1)$. 
Next, the set of instances with the same guessed label is constructed as $S_{\pos} =\{V'|V' \in \queue^{+}, \argmax(Y^{\queue^+}_j) = \argmax(\vec{Y}_i), \forall i \in [1, |x_b|], \forall j \in [1, N_{q+}]\}$, where $N_{q+}$ represents the length of $\queue+$. 
Following, the supervised contrastive loss is employed as: 
\begin{equation}\label{eq: supervised contrastive loss}\small
    \mathcal{L}_{s}^{ct}=-\frac{1}{|x_b|}\sum_{i=1}^{|x_b|} \{\frac{1}{|S_{\pos}|} \sum_{V' \in {S_{\pos}}}\log \frac{\exp(Q_i V'^\top /\tau)}{\sum_{V \in {\queue^{+}}} \exp(Q_i V^\top/\tau)}\}.
\end{equation}

Both the negative samples pool and the corresponding pseudo-label pool, are randomly initialized with Gaussian distributions. $\queue$ is replaced by the embeddings $K$ obtained in the previous training batch, and $Y^{\queue}$ is replaced by the uniformly distributed candidate labels $\overline{\hat{Y}}$, chronologically.

\textbf{Prototype-based Label Disambiguation.} Prototype is defined as a representative embedding of instances with the same label \cite{li2021prototypical, wang2022pico}. The prototype is denoted as $\proto \in \mathbb{R}^{k\times N_{\emb}}$, where the instances belonging to each class have their unique embedding vector. From an EM perspective, the E-step uses clustering to estimate the distribution of the prototype, and the M-step is to update the query network parameters through contrastive learning \cite{li2021prototypical, wang2022pico}. The prototype is zero-initialized and smoothly updated by the embedding $Q$ as: 
\begin{equation}\label{eq: prototype}
\begin{split}
    \proto_j &= ||\proto_j* \lambda + (1-\lambda)*Q_i   ||_2, \\
    &\text{if} ~ j = \argmax( \overline{\hat{Y}}\phi(p_m(x_i))), \forall i \in [1, |x_b|], \forall j \in [1, k],  
\end{split}
\end{equation} 
where $\lambda=0.99$ is the coefficient used in a moving average strategy. Following, the pseudo-label after prototype-based disambiguation is denoted as: 
\begin{equation}\label{eq: pseudo label}
    Y^{\proto}_i = \argmax(\phi(Q_i\proto^\top) \overline{\hat{Y_i}}).
\end{equation}
Finally, the total loss consists of the cross-entropy loss and the supervised contrastive loss using prototype-disambiguated labels, as
\begin{equation}\label{eq: PiCO}
    \mathcal{L} = \mathcal{L}_{ce} + \xi \mathcal{L}_s^{cont},
\end{equation} 
where $\vec{Y}$ refers to $Y^{\proto}$.
$\xi=0.5$ is adopted as the contrastive loss weight, as suggested in \cite{wang2022pico}. When the label disambiguation process is not used, the contrastive loss $\mathcal{L}_s^{ct}$ will be replaced by $\mathcal{L}_u^{ct}$, and $\vec{Y}$ will be replaced by $\hat{Y}$ in Eq. \ref{eq: PRODEN}. Note that when contrastive learning is not used, $\xi=0$.

\section{Experiment Setup} \label{sec:experiment}
\textcolor{black}{This section provides the details of the datasets used, followed by the descriptions of feature extraction, evaluation protocols, and model specifics, including the backbone, hyper-parameters, and training.
}

\subsection{Dataset}
\textcolor{black}{We use the SEED-IV\footnote{\href{https://bcmi.sjtu.edu.cn/home/seed/seed-iv.html}{https://bcmi.sjtu.edu.cn/home/seed/seed-iv.html}} \cite{zheng2018emotionmeter} and SEED-V \footnote{\href{https://bcmi.sjtu.edu.cn/home/seed/seed-v.html}{https://bcmi.sjtu.edu.cn/home/seed/seed-v.html}} \cite{liu2021comparing} datasets to evaluate and compare the different techniques described earlier.} 
\textcolor{black}{In SEED-IV, $72$ video clips with four emotions (`happy', `neutral', `sad', and `fear') served as stimuli. The study involved $15$ participants, comprised of $8$ females and $7$ males. Each subject repeated the experiment three times, each time with different stimuli. The experiment includes $24$ trials for each session, with $6$ trials per emotion. Each trial consists of three stages: $5$ seconds of start cue, a $2$-minute video clip, followed by a $45$-second period for self-assessment. A total of $62$ EEG recordings were collected at a sampling rate of $1000$\textit{Hz}.}
\textcolor{black}{In SEED-V, $45$ short films with five emotions (`happy', `neutral', `sad', `fear', and `disgust') were used as stimuli.} $16$ participants including $10$ females and $6$ males participated in the study. All participants repeated the experiment three times, with completely new stimuli each time. Each experiment contains $3$ trials for each emotion, yielding $15$ trials in total. Each trial includes three stages: $15$ seconds for initial stimuli prior to starting, $2-4$ minutes of watching film clips, and $15$ or $30$ seconds of self-assessment on the induction effect of the stimuli. EEG recordings were collected with $62$ channels at a sampling frequency of $1000$\textit{Hz}.

\textcolor{black}{\subsection{Candidate Label Generation}}
\subsubsection{Classical PLL Setup} \label{classical_setup} 
In the standard approach to EEG representation learning, one ground truth label is provided for each particular signal. However, in order to employ a PLL framework (classical), we can assume that the ground truth emotion label provided with the dataset is invariably part of the candidate label set. For each sample, every emotion label that does not match the ground truth is given an equal chance, determined by the predefined probability $q<1$, to be considered as a candidate label. This procedure is detailed in the problem setup section (Section 3.1). 

\subsubsection{\textcolor{black}{Circumplex PLL Setup}}\label{circumplex_setup}
\begin{figure}[t]
    \begin{center}
    \includegraphics[width=0.9\columnwidth]{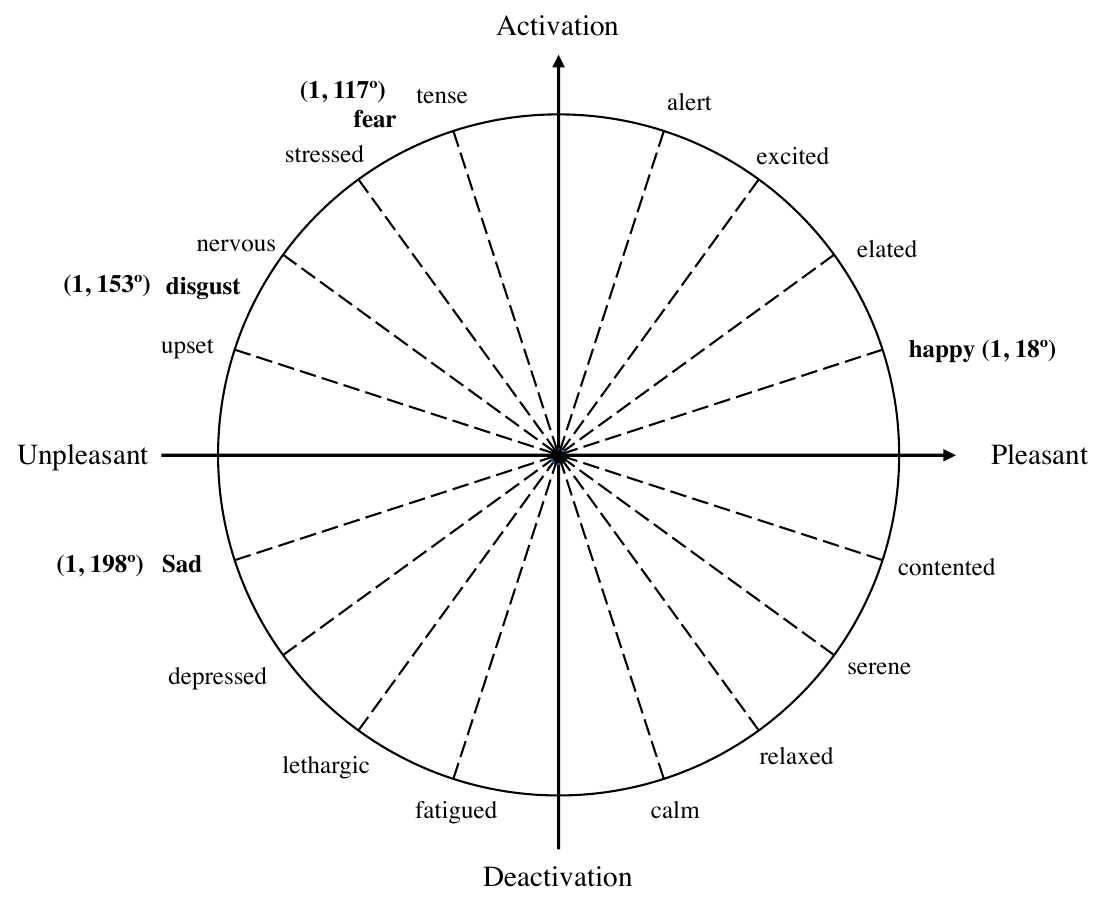} 
    \caption{Wheel of emotions reproduced from \cite{russell1980circumplex,zhong2019study}.}
        \label{emotion_wheel}
    \end{center}
\end{figure}

\begin{figure}
    \begin{center}
    \includegraphics[width=0.4\textwidth]{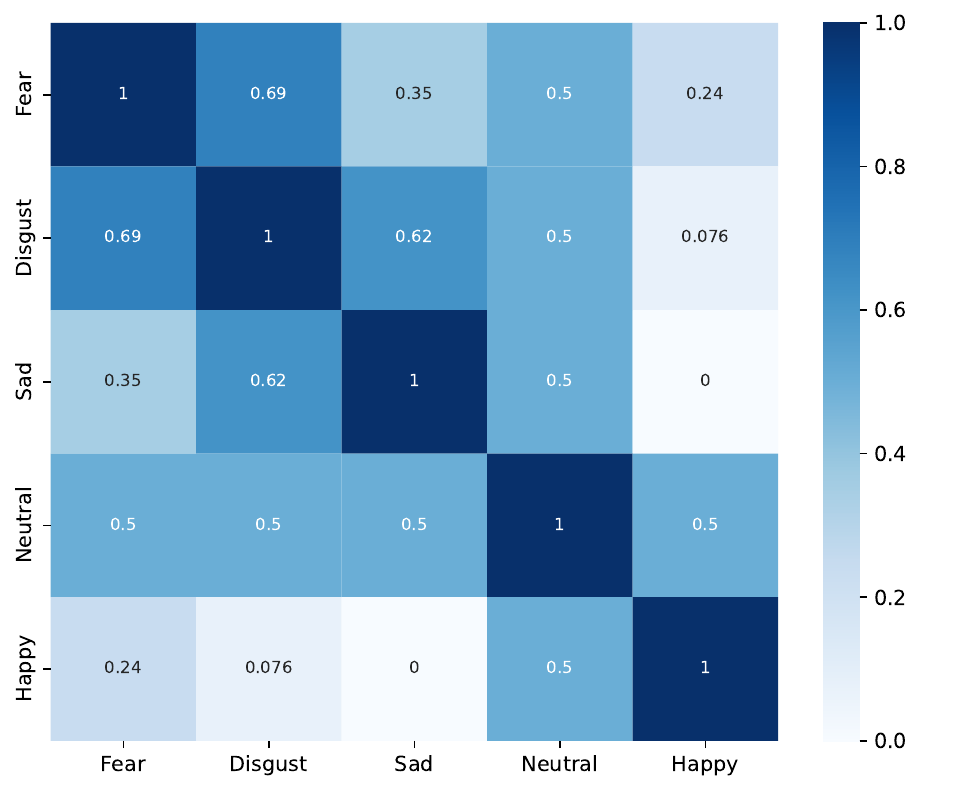} 
    \caption{\textcolor{black}{Normalized similarity score between two emotions as label ambiguity level in the circumplex PLL setup.}}
    \label{similarity}
    \end{center}
\end{figure}

\textcolor{black}{Most existing PLL methods employ independent sampling for candidate label generation, where each class except for the ground truth has the same probability of $q$ to become the candidate label, as mentioned in the classical PLL setup above. However, independent sampling may not be suitable for emotion recognition tasks since in Russell’s circumplex model \cite{russell1980circumplex}, it is more likely to confuse emotions that are more closely related (e.g., `sad' and `fear'), while it is less likely to confuse dissimilar emotions (e.g., `sad' and `happy'). To address this, we design additional experiments to generate the candidate labels based on the \textit{similarity} between two emotions. To do so, we first estimate the location of each emotion on Russell’s \textit{circumplex} model \cite{russell1980circumplex,zhong2019study}. As shown in Figure \ref{emotion_wheel}, the wheel of emotions is a circumplex that represents the relationships between different emotions on two orthogonal axes, arousal and valence. This was first proposed by Russell \cite{russell1980circumplex} and further developed in studies such as \cite{zhong2019study}. In this model, emotions are arranged in a circular layout, with certain emotions being considered more closely related to one another based on their proximity on the wheel. To perform more realistic experiments on label generation, we estimate the polar coordinates of each emotion in the format of (radius, angle degrees). According to \cite{zhong2019study}, we assume that emotions (with the exception of `disgust' and `fear') are uniformly distributed in each quarter of the wheel, with each adjacent angle representing a difference of $18^{\circ}$. `Disgust' is positioned between `upset' and `nervous', and `fear' is positioned between `stressed' and `tense'. Moreover, the `neutral' emotion is always placed at the center of the wheel, at the coordinate of $(0, 0^{\circ})$. Based on this distribution, we can determine the polar coordinates of each emotion, with the radius of the wheel being set to $1$. Following, we calculate the distance between two emotions ($i$ and $j$), as $\dist(i,j) =\sqrt{r_i^2 + r_j^2- 2 r_i r_j \cos{(\theta_i -\theta_j)}}$, where $r$ and $\theta$ denote the radius and angle radians, respectively. Next, we obtain the normalized similarity score between two emotions, as} 
\begin{equation} \label{gamma}
\gamma(i,j) =1- \dist(i,j)/ \max{(\dist(i,j))}, \forall i,j \in [1,k]. 
\end{equation}
\textcolor{black}{The normalized similarity scores among the five emotions are shown in Figure \ref{similarity}. Finally, for candidate label generation, we use the normalized similarity score instead of a predefined constant ($q$) as the probability, such that $\mathrm{P}(Y_i\mid x_i, y_i) = \prod_s \gamma(s,y_i), \forall s \in[1,k]$. Therefore, emotions that are more related to the ground truth have higher a probability of becoming the candidate labels and vice-versa.}


\subsubsection{\textcolor{black}{Real-World PLL Setup}}\label{real_world_setup}
\begin{figure}
    \begin{center}
    \includegraphics[width=1.0\columnwidth]{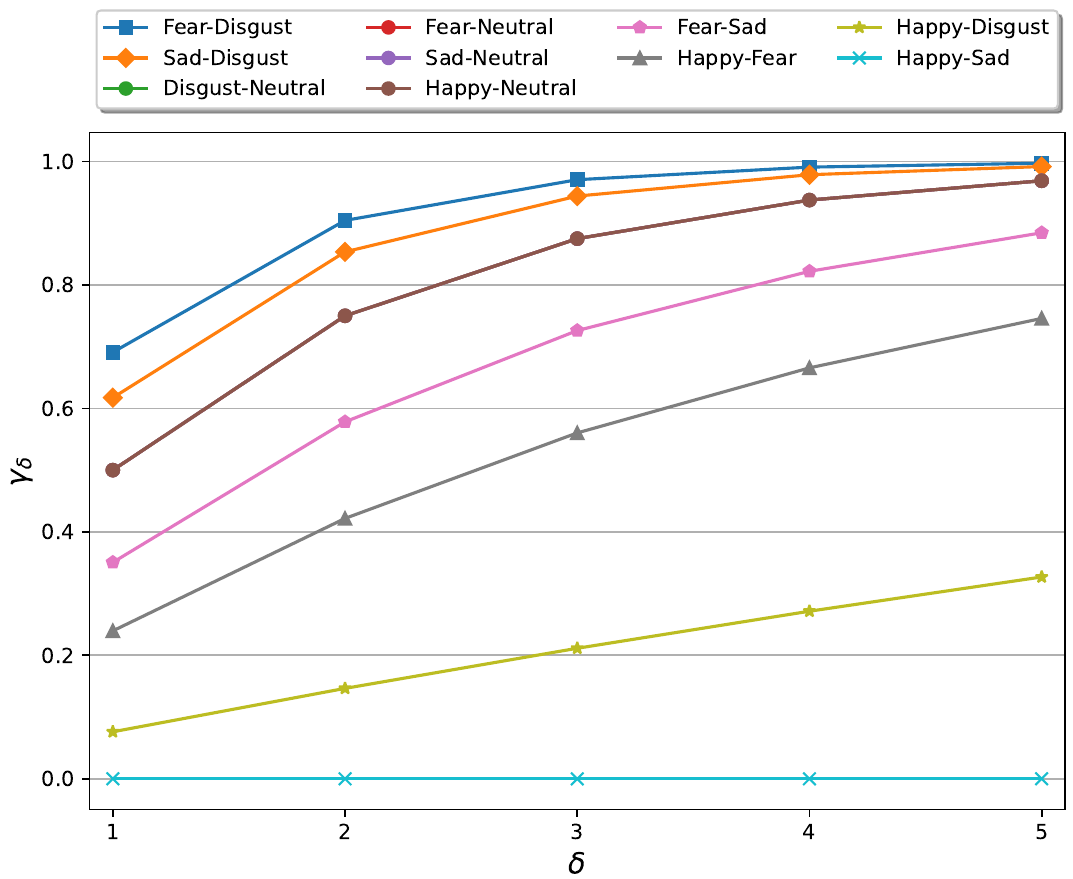} \color{black}
    \caption{Normalized similarity score ($\gamma_{\delta}$) between two emotions as label ambiguity level in the real-world PLL setup, with $\delta \in \{1, 2, 3, 4, 5\}$. The $\delta$ scaling controls the ambiguity level, affecting the similarity between emotion pairs based on their relative distances.}
    \label{emotion_pair_delta}
    \end{center}
\end{figure}

\textcolor{black}{Sections \ref{classical_setup} and \ref{circumplex_setup} describe the classical PLL setup with predefined and uniform label ambiguity levels as well as the circumplex PLL setup with non-uniform label ambiguity levels, respectively. 
Revisiting Russell’s circumplex model \cite{russell1980circumplex,zhong2019study}, we observe that the distances between emotions are predefined, offering a structured guide to emotional relationships. However, real-world emotional relationships are often more fluid, with emotions perceived in ways that may introduce greater ambiguity, potentially exceeding the model's structural limitations. To address this, we introduce {$\boldsymbol{\delta}$} \textbf{scaling}, a simple method to dynamically adjust the sensitivity of the circumplex model to emotional distances. We apply $\delta$ scaling to the normalized similarity score in Eq. \ref{gamma} as: 
\begin{equation}
\gamma_{\delta}(i,j) =1- [\dist(i,j)/ \max{(\dist(i,j))}]^{\delta}, \forall i,j \in [1,k]. 
\end{equation} 
Here, Eq. \ref{gamma} represents the special case of $\delta=1$.
Applying $\delta>1$ compresses most emotional distances, bringing them closer to one another. It should be noted that the similarity scores between identical emotions and very dissimilar emotions (e.g., `happy' vs. `sad') remain unchanged, as $\delta$ scaling does not affect the cases where normalized distances are $0$ or $1$. Therefore, with moderate $\delta$ values, $\delta$ scaling increases overall ambiguity by bringing similar emotions closer, while preserving the circumplex model's overall structure. Figure \ref{emotion_pair_delta} shows the normalized similarity scores ($\gamma_\delta$) of emotion pairs with $\delta \in \{1, 2, 3, 4, 5\}$. Among the $10$ emotion pairs, curves for `non-neutral'-`neutral' pairs (e.g., `sad'-`neutral', `fear'-`neutral') are identical. Generally, $\gamma_\delta$ increases as $\delta$ increases, except in the case of `happy'-`sad'.}

\textcolor{black}{We apply $\delta \in \{1, 2, 3, 4, 5\}$ in the real-world PLL setup to evaluate the performance of all PLL methods and assess the impact of disambiguation as average ambiguity increases. Furthermore, to enable fair performance comparison between classical and real-world PLL setups, we align the ambiguity level ($q$) in the classical PLL setup with the average ambiguity level calculated using similarity scores from the real-world setup, as follows:} 
\begin{equation}\label{average_q}\color{black}
q = \frac{\sum_{i=1}^{k} \sum_{j=i+1}^{k} \gamma_{\delta}(i,j)}{{k(k - 1)}/{2}}, \forall i,j \in [1,k].
\end{equation}

\subsection{Feature Space} \label{feature_space}
\begin{figure}[h!]\color{black}
    \begin{center}
    \includegraphics[width=1.0\columnwidth]{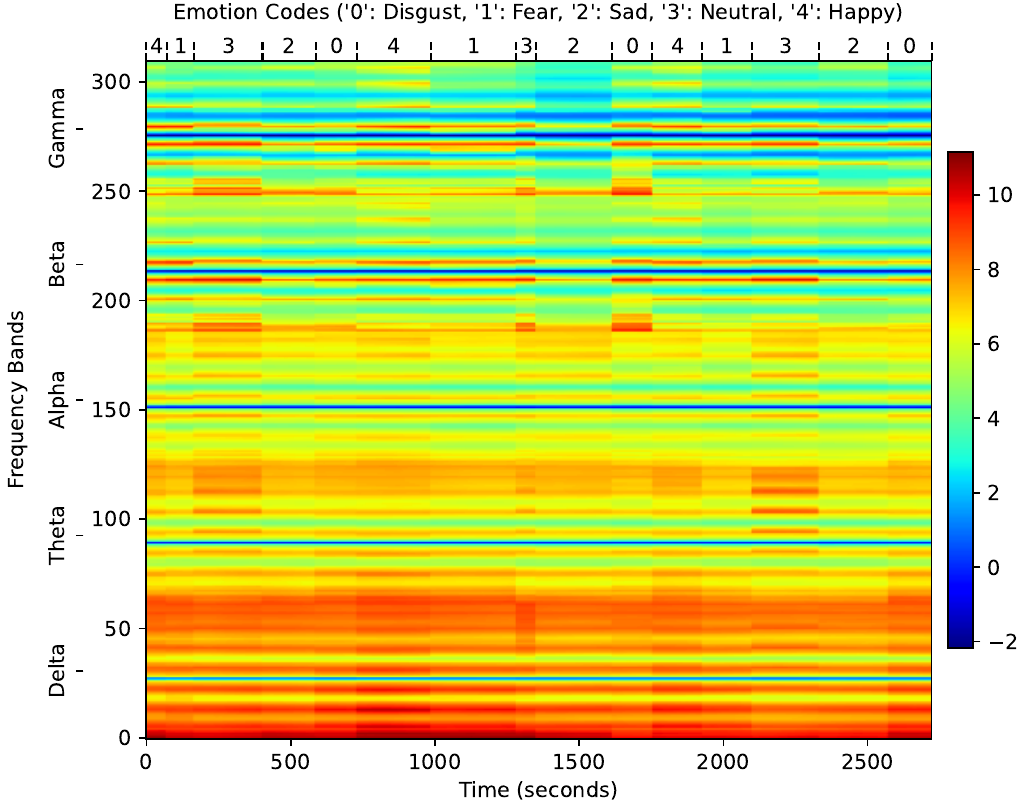} 
    \caption{DE feature maps from a single experiment of the SEED-V dataset ($5$ emotion categories) are provided. The bottom horizontal axis represents time, while the top horizontal axis displays emotion labels (0-4), with each label centered within dashed line boundaries to indicate the unique emotion category for this EEG segment. The vertical axis represents the $5$ EEG frequency bands and $62$ EEG channels in each band.}
    \label{visualization}
    \end{center}
\end{figure}

\begin{figure}[h!]\color{black}
    \begin{center}
    \includegraphics[width=1.0\columnwidth]{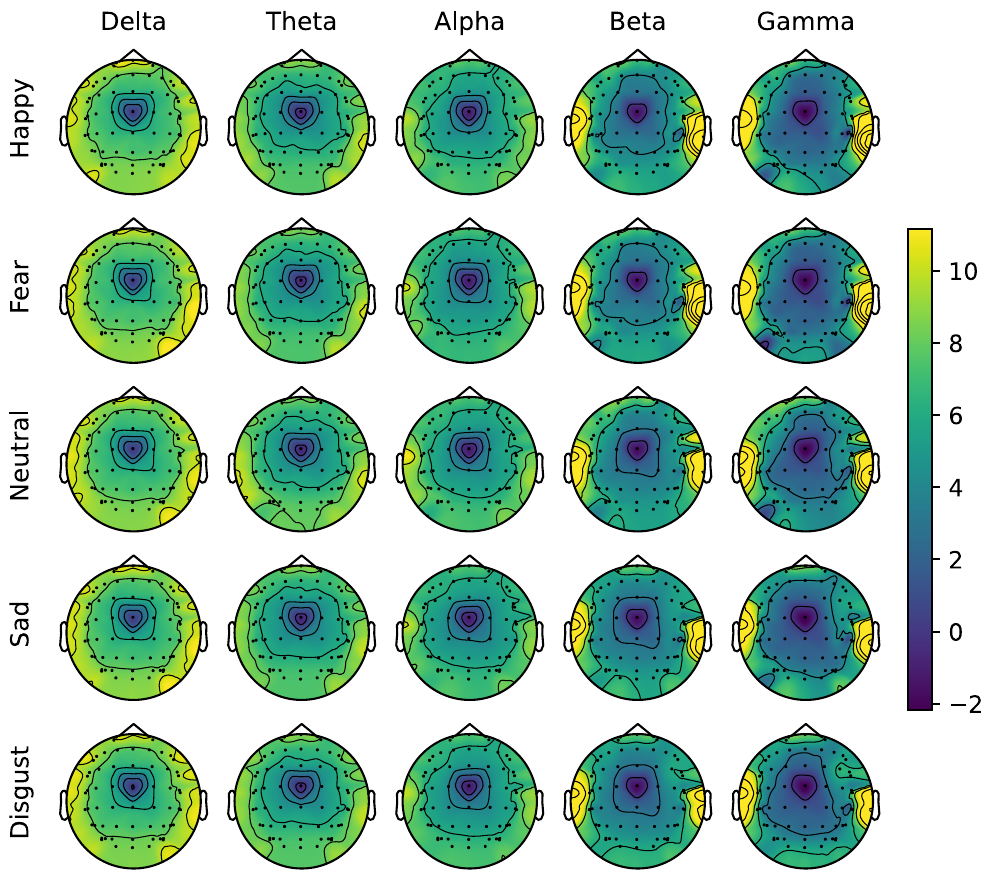} 
    \caption{Topographical maps of DE features from a single experiment in the SEED-V dataset are shown. The maps are averaged across all EEG segments for each emotion category (rows: Happy, Fear, Neutral, Sad, Disgust) and frequency band (columns: Delta, Theta, Alpha, Beta, Gamma). Each topographical plot illustrates the spatial distribution of DE values across $62$ EEG channels.}
    \label{topomap}
    \end{center}
\end{figure}

EEG recordings in each experiment were split into continuous and non-overlapping $4$-second segments. \textcolor{black}{We use the Differential Entropy (DE) features provided by the datasets \cite{zheng2018emotionmeter,liu2021comparing}.} DE features were extracted from $5$ EEG bands (delta, theta, alpha, beta, and gamma) and all $62$ EEG channels for each segment, yielding $s=310$ features in total. 
\textcolor{black}{Figure \ref{visualization} presents a thermogram of DE feature extracted from a single experiment in the SEED-V dataset \cite{liu2021comparing}. Each row corresponds to a specific EEG frequency band, concatenated across EEG channels, yielding a $1$-dimensional feature vector with $310$ features per $4$-second segment. The horizontal axis represents time in seconds, while the vertical axis shows the feature values across all bands and channels. Emotion codes are shown along the top of the figure, indicating the emotion label associated with each segment (`0': Disgust, `1': Fear, `2': Sad, `3': Neutral, `4': Happy).} 
\textcolor{black}{Additionally, Figure \ref{topomap} illustrates the spatial distribution of DE features across different emotion categories and frequency bands. These DE features are averaged over all EEG segments within the same single SEED-V experiment (as shown in Figure \ref{visualization}). We observe that DE values are consistently higher in the frontal and temporal lobes, particularly in higher-frequency bands such as beta and gamma. In contrast, regions such as the central, fronto-central, and centro-parietal lobes generally shows lower DE values across emotions. These patterns are consistent with findings in affective neuroscience where the frontal lobe, especially the prefrontal cortex, is critically involved in emotional regulation \cite{ochsner2005cognitive}, while the temporal lobe, plays a key role in emotional memory and reaction \cite{ledoux2013emotion}.}
We normalize the feature vector of each segment ($x_i \sim [0,1]^{310}$) to be used as input for training the models.

\subsection{Evaluation Protocols}
\textcolor{black}{In SEED-IV dataset, we follow the evaluation protocol defined in \cite{zheng2018emotionmeter}, where the first $16$ trials are used for training and the remaining $8$ are used for testing.  
In SEED-V dataset, we apply the same evaluation protocol that was originally used in \cite{liu2021comparing}.} Specifically, EEG recordings of each subject have been formed into three predefined folds. Fold No. 1 includes the first $5$ trials of all three repeated experiments, yielding $15$ trials in total. A similar strategy is applied to folds No. $2$ and $3$. Next, a 3-fold cross-validation scheme has been used for training and testing, as suggested in \cite{liu2021comparing}.

\subsection{Backbone Model}\label{backbone}
\textcolor{black}{We employ the same lightweight CNN used in very recent EEG-based affective computing studies \cite{zhang2022holistic,zhang2022parse} as our backbone model due to its demonstrated effectiveness. The lightweight nature of this CNN makes it particularly well-suited for handling EEG data, balancing model efficiency with the ability to capture important spatial and temporal features. This structure is well-suited for EEG signals, as they are high-dimensional but lack the large labeled datasets found in other domains such as computer vision and natural language processing, allowing for faster training and inference without the need for a deeper network.} As shown in Table \ref{model}, the encoder consists of two 1-dimensional (1-D) convolutional blocks. Each block has a 1-D convolutional layer followed by a 1-D batch normalization layer and a LeakyReLU activation function. \textcolor{black}{The encoder is used to transform the DE features extracted from EEG recordings into a compact representation.} The learned embedding is then fed to the classifier which includes two fully connected layers with a dropout rate of 0.5 for emotion recognition. In the table, $s$ and $k$ denote the total number of EEG features, and the number of emotion categories, respectively.

\begin{table}[t]
\caption{Backbone Model Details.}
\scriptsize
\label{model}
\resizebox{\columnwidth}{!}
{
\begin{tabular}{c|c|c}
\hline
\textbf{Module} & \textbf{Layer details}  & \textbf{Output shape} \\[.1cm] \hline
\textbf{Input} & -  & $(1, s)$ \\[.1cm] \hline
\multirow{4}{*}{\textbf{Encoder}} & \convMatrix{3}{5}{5} & $(5, s-2)$ \\[.1cm] 
\cline{2-3} & \convMatrix{3}{10}{10} & $(10, s-4)$ \\[.1cm]
\cline{2-3} \hline \textbf{Embedding} & Flatten & $10\times (s-4)$ \\[.1cm]
\cline{2-3}
\hline
\multirow{2}{*}{\textbf{Classifier}} & \convMatrixc{k} & $(k)$ \\[.1cm] 
\hline

\end{tabular}
}
\end{table}

\subsection{Implementation Details}
In all the experiments, we ran $T=30$ training epochs with a batch size of $|x_b|=8$. We set the learning rate to $0.01$ and used SGD as the optimizer with a default momentum of $0.9$ and a weight decay of $0.0001$.
The learning rate scheduler was not used for the naive method \cite{seo2021power} and the fully supervised method as it hurt the performance in these cases. All the PLL methods have been evaluated \textbf{five} times, each time with different random seeds used for candidate label generation.
Our experiments were carried out on two NVIDIA GeForce RTX $2080$ Ti GPUs using PyTorch \cite{paszke2019pytorch}.
For the sake of reproducibility, we made the source code of this work publicly available at: \href{https://github.com/guangyizhangbci/PLL-Emotion-EEG}{https://github.com/guangyizhangbci/PLL-Emotion-EEG}.

\begin{table*}[!ht]\color{black}
    \centering
    \setlength
    \tabcolsep{10pt}
    \caption{The accuracy (in \%) of Comparison PLL Methods (SEED-IV). \textcolor{black}{The average accuracy across $15$ subjects and $3$ repeated sessions is presented, with standard deviations shown in parentheses.}}
    \begin{tabular}{c|c|c|cccccc}
    \toprule
    Method          &Venue & LD                                               & $q=0.2$               & $q=0.4$              & $q=0.6$              & $q=0.8$              & $q=0.9$             & $q=0.95$             \\
        	\midrule

  \cellcolor{Gray} Fully Supervised &\cellcolor{Gray}-&\cellcolor{Gray}-& \multicolumn{6}{c}{\cellcolor{Gray}62.11 \scriptsize(16.58)}  \\

	\midrule    

    \multirow{2}*{PRODEN \cite{lv2020progressive}}	& \multirow{2}*{ICML20}  &\xmark    &58.23\scriptsize(17.98)	&	57.44\scriptsize(16.66)	&	52.85\scriptsize(18.41)	&	47.18\scriptsize(16.32)	&	39.77\scriptsize(15.93)	&	32.46\scriptsize(16.35)	\\	
    &                          &\checkmark                                              &59.05\scriptsize(17.45)	&	58.77\scriptsize(16.42)	&	56.86\scriptsize(17.35)	&	54.53\scriptsize(15.39)	&	46.07\scriptsize(15.38)	&	\underline{38.76\scriptsize(15.76)}	\\
	\midrule    

    DNPL    \cite{seo2021power}         & ICASSP21  & \xmark                            &58.38\scriptsize(17.51)	&	\underline{58.95\scriptsize(18.24)}	&	\textbf{59.18\scriptsize(18.29)}	&	\textbf{59.06\scriptsize(17.64)}	&	\textbf{57.02\scriptsize(17.99)}	&	\textbf{55.49\scriptsize(17.32)}	\\	

	\midrule

    \multirow{2}*{LW \cite{wen2021leveraged}}	& \multirow{2}*{ICML21}      &\xmark    &\underline{59.31\scriptsize(17.18)}	&	58.56\scriptsize(17.45)	&	56.13\scriptsize(17.52)	&	51.88\scriptsize(16.11)	&	44.50\scriptsize(15.63)	&35.75\scriptsize(17.46)	\\
    &                         &\checkmark                                               &\textbf{59.54\scriptsize(17.13)}	&	\textbf{59.34\scriptsize(16.63)}	&	\underline{58.35\scriptsize(17.21)}	&	\underline{58.90\scriptsize(17.60)}	&	\underline{51.68\scriptsize(17.68)}	&	37.67\scriptsize(19.62)	\\	
    
	\midrule

    \multirow{2}*{CAVL \cite{zhang2022exploiting}}	& \multirow{2}*{ICLR22}  &\xmark    &58.26\scriptsize(17.99)	&	57.43\scriptsize(16.66)	&	52.85\scriptsize(18.40)	&	47.19\scriptsize(16.34)	&	39.76\scriptsize(15.93)	&	32.46\scriptsize(16.35)	\\
    &                         &\checkmark                                               &59.13\scriptsize(17.38)	&	58.36\scriptsize(16.81)	&	55.46\scriptsize(17.02)	&	47.64\scriptsize(16.90)	&	36.58\scriptsize(14.67)	&	30.35\scriptsize(12.03)	\\
    
	\midrule    
    \multirow{2}*{CR \cite{wu2022revisiting}}	& \multirow{2}*{ICML22}      &\xmark    &50.45\scriptsize(18.01)	&	49.96\scriptsize(18.05)	&	48.34\scriptsize(17.49)	&	45.68\scriptsize(15.86)	&	39.91\scriptsize(14.89)	&	36.21\scriptsize(14.91)	\\	
    &                         &\checkmark                                               &42.91\scriptsize(16.23)	&	39.72\scriptsize(14.57)	&	37.07\scriptsize(14.82)	&	32.85\scriptsize(14.08)	&	29.30\scriptsize(11.53)	&	27.00\scriptsize(11.33)	\\

	\midrule        
    
    \multirow{2}*{PiCO \cite{wang2022pico}}	& \multirow{2}*{ICLR22}          &\xmark    &56.17\scriptsize(17.58)	&	53.95\scriptsize(17.48)	&	52.03\scriptsize(18.73)	&	42.25\scriptsize(17.89)	&	34.81\scriptsize(17.71)	&	30.51\scriptsize(16.78)	\\
    &                         &\checkmark                                               &58.31\scriptsize(17.51)	&	58.40\scriptsize(16.89)	&	56.81\scriptsize(17.72)	&	50.69\scriptsize(17.35)	&	42.23\scriptsize(16.65)	&	36.13\scriptsize(18.81)	\\

    \bottomrule
    \end{tabular}
    \label{table: comparison_seed-iv}
\end{table*}

\begin{table*}[!ht]
    \centering
    \setlength
    \tabcolsep{10pt}
    \caption{The accuracy (in \%) of Comparison PLL Methods (SEED-V). \textcolor{black}{The average accuracy across $16$ subjects and $3$ repeated sessions is presented, with standard deviations shown in parentheses.}}
    \begin{tabular}{c|c|c|cccccc}
    \toprule
    Method          &Venue & LD                                               & $q=0.2$               & $q=0.4$              & $q=0.6$              & $q=0.8$              & $q=0.9$             & $q=0.95$             \\
        	\midrule

  \cellcolor{Gray} Fully Supervised &\cellcolor{Gray}-&\cellcolor{Gray}-& \multicolumn{6}{c}{\cellcolor{Gray}63.08 \scriptsize(13.87)}  \\

	\midrule    

    \multirow{2}*{PRODEN \cite{lv2020progressive}}	& \multirow{2}*{ICML20}  &\xmark    &58.55\scriptsize(16.63)	&	57.69\scriptsize(15.99)	&	53.87\scriptsize(15.24)	&	43.05\scriptsize(14.76)	&	32.37\scriptsize(11.73)	&	26.00\scriptsize(10.72)	\\
    &                          &\checkmark                                              &59.73\scriptsize(16.81)	&	58.83\scriptsize(16.12)	&	55.87\scriptsize(16.37)	&	47.53\scriptsize(14.06)	&	\underline{37.58\scriptsize(13.54)}	&	26.99\scriptsize(10.32)	\\
	\midrule    

    DNPL    \cite{seo2021power}         & ICASSP21  & \xmark                            &\underline{60.86\scriptsize(16.64)}	&	60.37\scriptsize(15.82)	&	\underline{59.62\scriptsize(16.45)}	&	\textbf{59.42\scriptsize(15.65)}	&	\textbf{57.08\scriptsize(16.46)}	&	\textbf{49.44\scriptsize(15.85)}	\\

	\midrule

    \multirow{2}*{LW \cite{wen2021leveraged}}	& \multirow{2}*{ICML21}      &\xmark    &60.64\scriptsize(15.83)	&	58.86\scriptsize(16.03)	&	55.97\scriptsize(15.91)	&	48.43\scriptsize(14.00)	&	36.09\scriptsize(12.14)	&	\underline{29.30\scriptsize(11.57)}	\\
    &                         &\checkmark                                               &60.48\scriptsize(16.75)	&	\underline{60.83\scriptsize(16.22)}	&	\textbf{59.81\scriptsize(16.30)}	&	\underline{54.61\scriptsize(16.30)}	&	34.35\scriptsize(17.91)	&	22.12\scriptsize(10.33)	\\
    
	\midrule

    \multirow{2}*{CAVL \cite{zhang2022exploiting}}	& \multirow{2}*{ICLR22}  &\xmark    &58.51\scriptsize(16.63)	&	57.60\scriptsize(15.90)	&	53.67\scriptsize(15.24)	&	43.03\scriptsize(14.70)	&	32.39\scriptsize(11.66)	&	26.03\scriptsize(10.68)	\\
    &                         &\checkmark                                               &57.58\scriptsize(16.52)	&	54.94\scriptsize(16.68)	&	48.58\scriptsize(15.64)	&	30.75\scriptsize(12.18)	&	23.44 $ $\scriptsize(8.19)	&	21.72 $ $\scriptsize(7.03)	\\	
    
	\midrule    
    \multirow{2}*{CR \cite{wu2022revisiting}}	& \multirow{2}*{ICML22}      &\xmark    &42.42\scriptsize(13.70)	&	42.64\scriptsize(14.37)	&	41.22\scriptsize(13.56)	&	35.60\scriptsize(12.62)	&	30.70\scriptsize(10.71)	&	26.41 $ $\scriptsize(9.27)	\\
    &                         &\checkmark                                               &28.31\scriptsize(10.08)	&	28.55 $ $\scriptsize(9.52)	&	27.57 $ $\scriptsize(9.57)	&	22.68 $ $\scriptsize(8.66)	&	22.09 $ $\scriptsize(7.90)	&	20.94 $ $\scriptsize(8.56)	\\	

	\midrule        
    
    \multirow{2}*{PiCO \cite{wang2022pico}}	& \multirow{2}*{ICLR22}          &\xmark    &57.23\scriptsize(16.44)	&	55.28\scriptsize(16.71)	&	49.95\scriptsize(15.89)	&	38.04\scriptsize(12.82)	&	28.34\scriptsize(11.18)	&	24.91\scriptsize(10.26)	\\
    &                         &\checkmark                                               &\textbf{62.68\scriptsize(15.88)}	&	\textbf{61.92\scriptsize(15.94)}	&	57.54\scriptsize(15.59)	&	43.87\scriptsize(14.99)	&	31.26\scriptsize(11.52)	&	24.69 $ $\scriptsize(9.66)	\\	

    \bottomrule
    \end{tabular}
    \label{table: comparison_seed-v}
\end{table*}

\begin{table*}[!ht]\color{black}
    \centering
    \setlength
    \tabcolsep{10pt}
    \caption{The accuracy (in \%) of LW \cite{wen2021leveraged} with different settings (SEED-IV). \textcolor{black}{The average accuracy across $15$ subjects and $3$ repeated sessions is presented, with standard deviations shown in parentheses.}}
    \begin{tabular}{c|c|c|cccccc}
    \toprule
        Method          & $\beta$ & LD                    & $q=0.2$               & $q=0.4$              & $q=0.6$              & $q=0.8$              & $q=0.9$             & $q=0.95$             \\
        	\midrule
           
        \multirow{6}*{LW-Sigmoid}    &  \multirow{2}*{0} &\xmark    &48.40\scriptsize(18.05)	&	44.03\scriptsize(17.46)	&	37.72\scriptsize(17.04)	&	32.06\scriptsize(15.15)	&	28.04\scriptsize(15.62)	&	26.77\scriptsize(13.40)	\\
                                     &  &\checkmark                 &43.54\scriptsize(17.27)	&	39.67\scriptsize(15.72)	&	33.75\scriptsize(13.86)	&	29.96\scriptsize(11.83)	&	26.99\scriptsize(10.64)	&	25.24 $ $\scriptsize(9.99)	\\
                                     \cmidrule{2-9}
                                     &  \multirow{2}*{1} &\xmark    &57.71\scriptsize(17.87)	&	49.31\scriptsize(18.30)	&	44.73\scriptsize(16.90)	&	38.28\scriptsize(15.93)	&	31.25\scriptsize(16.14)	&	28.01\scriptsize(13.96)	\\
                                     &  &\checkmark                 &50.01\scriptsize(20.17)	&	45.32\scriptsize(19.63)	&	34.47\scriptsize(17.73)	&	30.14\scriptsize(12.97)	&	27.89\scriptsize(12.59)	&	26.25 $ $\scriptsize(9.93)	\\
                                     \cmidrule{2-9}
                                     &  \multirow{2}*{2} &\xmark    &58.78\scriptsize(16.90)	&	57.82\scriptsize(17.64)	&	49.56\scriptsize(17.34)	&	41.02\scriptsize(16.61)	&	34.41\scriptsize(16.73)	&	28.80\scriptsize(14.55)	\\	
                                     &  &\checkmark                 &57.41\scriptsize(19.42)	&	54.06\scriptsize(20.06)	&	46.11\scriptsize(21.44)	&	32.12\scriptsize(15.54)	&	27.71\scriptsize(14.00)	&	26.41\scriptsize(10.95)	\\
        \midrule
    \multirow{6}*{LW-Cross Entropy}  & \multirow{2}*{0}  &\xmark    &57.51\scriptsize(18.46)	&	56.92\scriptsize(17.13)	&	53.79\scriptsize(17.69)	&	46.10\scriptsize(17.41)	&	38.55\scriptsize(15.61)	&	31.68\scriptsize(16.58)	\\
                                     &  &\checkmark                 &59.05\scriptsize(17.45)	&	58.78\scriptsize(16.43)	&	56.87\scriptsize(17.33)	&	54.52\scriptsize(15.39)	&	46.08\scriptsize(15.37)	&	\textbf{38.76\scriptsize(15.76)}	\\	
                                     \cmidrule{2-9}
                                     & \multirow{2}*{1}  &\xmark    &58.36\scriptsize(17.77)	&	57.71\scriptsize(17.31)	&	55.26\scriptsize(17.51)	&	49.27\scriptsize(16.93)	&	41.87\scriptsize(15.39)	&	34.08\scriptsize(16.91)	\\
                                     &  &\checkmark                 &\underline{59.39\scriptsize(16.96)}	&	\underline{59.08\scriptsize(16.51)}	&	\underline{58.08\scriptsize(17.33)}	&	\underline{57.50\scriptsize(16.82)}	&	\underline{49.60\scriptsize(17.45)}	&	36.60\scriptsize(18.26)	\\	
                                     \cmidrule{2-9}
                                     & \multirow{2}*{2}  &\xmark    &59.31\scriptsize(17.18)	&	58.56\scriptsize(17.45)	&	56.13\scriptsize(17.52)	&	51.88\scriptsize(16.11)	&	44.50\scriptsize(15.63)	&	35.75\scriptsize(17.46)	\\	
                                     &  &\checkmark                 &\textbf{59.54\scriptsize(17.13)}	&	\textbf{59.34\scriptsize(16.63)}&	\textbf{58.35\scriptsize(17.21)}	&	\textbf{58.90\scriptsize(17.60)}	&	\textbf{51.68\scriptsize(17.68)}	&	\underline{37.67\scriptsize(19.62)}	\\

    \bottomrule
    \end{tabular}
    \label{table: LW (seed-iv)}
\end{table*}

\begin{table*}[!ht]
    \centering
    \setlength
    \tabcolsep{10pt}
    \caption{The accuracy (in \%) of LW \cite{wen2021leveraged} with different settings (SEED-V). \textcolor{black}{The average accuracy across $16$ subjects and $3$ repeated sessions is presented, with standard deviations shown in parentheses.}}
    \begin{tabular}{c|c|c|cccccc}
    \toprule
        Method          & $\beta$ & LD                    & $q=0.2$               & $q=0.4$              & $q=0.6$              & $q=0.8$              & $q=0.9$             & $q=0.95$             \\
        	\midrule
           
        \multirow{6}*{LW-Sigmoid}    &  \multirow{2}*{0} &\xmark    &35.68\scriptsize(12.69)	&	31.25\scriptsize(11.01)	&	26.79\scriptsize(10.01)	&	22.14 $ $\scriptsize(9.22)	&	22.47 $ $\scriptsize(9.26)	&	20.66 $ $\scriptsize(8.84)	\\
                                     &  &\checkmark                 &30.14\scriptsize(10.52)	&	27.88\scriptsize(10.03)	&	23.03 $ $\scriptsize(8.03)	&	20.98 $ $\scriptsize(6.79)	&	20.11 $ $\scriptsize(6.52)	&	20.00 $ $\scriptsize(6.65)	\\
                                     \cmidrule{2-9}
                                     &  \multirow{2}*{1} &\xmark    &57.44\scriptsize(16.66)	&	48.55\scriptsize(15.32)	&	34.54\scriptsize(11.24)	&	25.77 $ $\scriptsize(9.85)	&	24.35 $ $\scriptsize(9.68)	&	22.08 $ $\scriptsize(9.62)	\\
                                     &  &\checkmark                 &40.99\scriptsize(23.27)	&	33.97\scriptsize(19.67)	&	24.74\scriptsize(10.24)	&	20.75 $ $\scriptsize(7.80)	&	20.04 $ $\scriptsize(6.99)	&	20.61 $ $\scriptsize(6.72)	\\
                                     \cmidrule{2-9}
                                     &  \multirow{2}*{2} &\xmark    &36.72\scriptsize(15.55)	&	48.96\scriptsize(18.35)	&	49.51\scriptsize(15.64)	&	28.98\scriptsize(11.12)	&	24.54 $ $\scriptsize(9.70)	&	22.52 $ $\scriptsize(9.40)	\\
                                     &  &\checkmark                 &42.44\scriptsize(23.89)	&	37.82\scriptsize(22.15)	&	32.85\scriptsize(19.30)	&	21.56 $ $\scriptsize(9.26)	&	20.25 $ $\scriptsize(6.91)	&	20.53 $ $\scriptsize(7.35)	\\	
        \midrule
    \multirow{6}*{LW-Cross Entropy}  & \multirow{2}*{0}  &\xmark    &59.44\scriptsize(16.16)	&	57.17\scriptsize(16.03)	&	53.93\scriptsize(15.14)	&	42.63\scriptsize(14.17)	&	31.81\scriptsize(12.40)	&	27.29\scriptsize(11.04)	\\
                                     &  &\checkmark                 &59.71\scriptsize(16.81)	&	58.85\scriptsize(16.08)	&	55.73\scriptsize(16.23)	&	48.13\scriptsize(13.88)	&	\textbf{36.83\scriptsize(13.21)}	&27.81\scriptsize(10.48)	\\
                                     \cmidrule{2-9}
                                     & \multirow{2}*{1}  &\xmark    &60.14\scriptsize(15.85)	&	58.49\scriptsize(16.13)	&	55.58\scriptsize(15.96)	&	46.38\scriptsize(14.55)	&	33.95\scriptsize(12.81)	&	\underline{28.27\scriptsize(11.14)}	\\
                                     &  &\checkmark                 &60.23\scriptsize(16.84)	&	\underline{60.01\scriptsize(16.36)}	&	\underline{58.79\scriptsize(16.14)}	&	\underline{51.48\scriptsize(16.19)}	&	30.47\scriptsize(15.87)	&	22.05\scriptsize(10.25)	\\
                                     \cmidrule{2-9}
                                     & \multirow{2}*{2}  &\xmark    &\textbf{60.64\scriptsize(15.83)}	&	58.86\scriptsize(16.03)	&	55.97\scriptsize(15.91)	&	48.43\scriptsize(14.00)	&	\underline{36.09\scriptsize(12.14)}	&	\textbf{29.30\scriptsize(11.57)}	\\	
                                     &  &\checkmark                 &\underline{60.48\scriptsize(16.75)}	&	\textbf{60.83\scriptsize(16.22)}	&	\textbf{59.81\scriptsize(16.30)}	&	\textbf{54.61\scriptsize(16.30)}	&	34.35\scriptsize(17.91)	&	22.12\scriptsize(10.33)	\\

    \bottomrule
    \end{tabular}
    \label{table: LW (seed-v)}
\end{table*}

\begin{table*}[!ht]\color{black}
    \centering
    \setlength
    \tabcolsep{10pt}
    \caption{The accuracy (in \%) of PiCO \cite{wang2022pico} with different settings (SEED-IV). \textcolor{black}{The average accuracy across $15$ subjects and $3$ repeated sessions is presented, with standard deviations shown in parentheses.}}
    \begin{tabular}{c|c|cccccc}
    \toprule
         CL  & LD                     & $q=0.2$               & $q=0.4$              & $q=0.6$              & $q=0.8$              & $q=0.9$             & $q=0.95$             \\
        \midrule
        \xmark & \xmark                 &\underline{58.40\scriptsize(18.27)}	&	55.37\scriptsize(17.82)	&	54.00\scriptsize(17.09)	&	45.29\scriptsize(16.58)	&	39.41\scriptsize(16.93)	&	32.31\scriptsize(15.00)	\\
        \xmark &\checkmark              &\textbf{59.68\scriptsize(17.65)}	&	\underline{58.16\scriptsize(17.74)}	&	\textbf{58.55\scriptsize(17.37)}	&	\textbf{54.12\scriptsize(17.65)}	&	\textbf{49.17\scriptsize(18.45)}	&	\textbf{41.49\scriptsize(18.48)}	\\
        \checkmark       &   \xmark     &56.17\scriptsize(17.58)	&	53.95\scriptsize(17.48)	&	52.03\scriptsize(18.73)	&	42.25\scriptsize(17.89)	&	34.81\scriptsize(17.71)	&	30.51\scriptsize(16.78)	\\
        \checkmark & \checkmark         &58.31\scriptsize(17.51)	&	\textbf{58.40\scriptsize(16.89)}	&	\underline{56.81\scriptsize(17.72)}	&	\underline{50.69\scriptsize(17.35)}	&	\underline{42.23\scriptsize(16.65)}	&	\underline{36.13\scriptsize(18.81)}	\\

    \bottomrule
    \end{tabular}
    \label{table: PiCO (seed-iv)}
\end{table*}

\begin{table*}[!ht]
    \centering
    \setlength
    \tabcolsep{10pt}
    \caption{The accuracy (in \%) of PiCO \cite{wang2022pico} with different settings (SEED-V). \textcolor{black}{The average accuracy across $16$ subjects and $3$ repeated sessions is presented, with standard deviations shown in parentheses.}}
    \begin{tabular}{c|c|cccccc}
    \toprule
         CL  & LD                     & $q=0.2$               & $q=0.4$              & $q=0.6$              & $q=0.8$              & $q=0.9$             & $q=0.95$             \\
        \midrule
        \xmark & \xmark                 &59.35\scriptsize(16.09)	&	57.75\scriptsize(16.27)	&	53.13\scriptsize(15.84)	&	43.31\scriptsize(13.41)	&	\underline{33.30\scriptsize(12.53)}	&	\textbf{27.12\scriptsize(10.52)}	\\	
        \xmark &\checkmark              &\underline{59.48\scriptsize(16.27)}	&	\underline{58.77\scriptsize(16.65)}	&	\underline{55.21\scriptsize(16.52)}	&	\textbf{45.93\scriptsize(16.42)}	&	\textbf{35.12\scriptsize(13.51)}	&	\underline{25.81 $ $\scriptsize(9.60)}	\\
        \checkmark       &   \xmark     &57.23\scriptsize(16.44)	&	55.28\scriptsize(16.71)	&	49.95\scriptsize(15.89)	&	38.04\scriptsize(12.82)	&	28.34\scriptsize(11.18)	&	24.91\scriptsize(10.26)	\\
        \checkmark & \checkmark         &\textbf{62.68\scriptsize(15.88)}	&	\textbf{61.92\scriptsize(15.94)}	&	\textbf{57.54\scriptsize(15.59)}	&	\underline{43.87\scriptsize(14.99)}	&	31.26\scriptsize(11.52)	&	24.69 $ $\scriptsize(9.66)	\\

    \bottomrule
    \end{tabular}
    \label{table: PiCO (seed-v)}
\end{table*}

\section{Results and Discussions} \label{sec:result}
\textcolor{black}{This section presents the performance of the PLL methods for EEG-based emotion recognition in classical and circumplex PLL setups, as well as in the real-world scenario designed based on the proximity of certain emotion classes. This section also discusses the impact of ground truth reliability.
}

\subsection{Performance in the Classical PLL Setup}
\textcolor{black}{We evaluate the performance of each PLL method in all $6$ scenarios with different levels of ambiguity ($q\in \{0.2, 0.4, 0.6, 0.8, 0.9, 0.95 \}$) and present the results in Table \ref{table: comparison_seed-iv} for SEED-IV and Table \ref{table: comparison_seed-v} for SEED-V. We observe that when candidate labels with low ambiguity ($q \in \{0.2, 0.4\}$) are provided, LW \cite{wen2021leveraged} and PiCO \cite{wang2022pico} obtain the best results in SEED-IV and SEED-V, respectively (shown in bold). DNPL \cite{seo2021power} obtains the second-best performance (shown with an underline) with less ambiguous labels ($q=0.4$ in SEED-IV and $q=0.2$ in SEED-V) while achieving the best performance with more ambiguous labels ($q\in \{0.8, 0.9, 0.95 \}$) in both the datasets. When the label ambiguity is moderate, DNPL \cite{seo2021power} and LW \cite{wen2021leveraged} behave similarly. Specifically, when $q=0.6$, DNPL ranks first in SEED-IV and second in SEED-V, while LW performs the best in SEED-V and second best in SEED-IV. Overall, across all the scenarios, DNPL \cite{seo2021power} maintains a very stable performance on both datasets, while others suffer from major performance drops when candidate labels are provided with very high ambiguity ($q > 0.8$).}

All PLL methods use label disambiguation methods \textit{except} for DNPL \cite{seo2021power}. To analyze the effect of label disambiguation in each PLL method, we evaluate them both with and without label disambiguation. \textcolor{black}{As shown in Tables \ref{table: comparison_seed-iv} and \ref{table: comparison_seed-v}, among these five PLL methods, label disambiguation methods, denoted by LD in the tables, play different roles. Specifically, label disambiguation consistently improves the model performance in PRODEN \cite{lv2020progressive} and PiCO \cite{wang2022pico} (only except when $q=0.95$ is used in SEED-V), while causing performance decline in CR \cite{wu2022revisiting}, across all the ambiguously labeled scenarios, on both datasets. In SEED-IV, for the CAVL method \cite{zhang2022exploiting}, label disambiguation improves the model performance when candidate labels with lower ambiguity are provided ($q \in \{0.4, 0.6, 0.8\}$). However, when candidate labels are provided with very high ambiguity ($q \in \{0.9, 0.95 \}$), label disambiguation results in performance degradation. In SEED-V, the LW method \cite{wen2021leveraged} exhibits a similar impact for label disambiguation on performance as seen with the CAVL method in SEED-IV. Notably, label disambiguation causes a consistent performance drop for CAVL in SEED-V while consistently increasing performance for LW  in SEED-IV dataset.}

\begin{figure}[htp!]\color{black}
    \begin{center}
    \includegraphics[width=0.8\columnwidth]{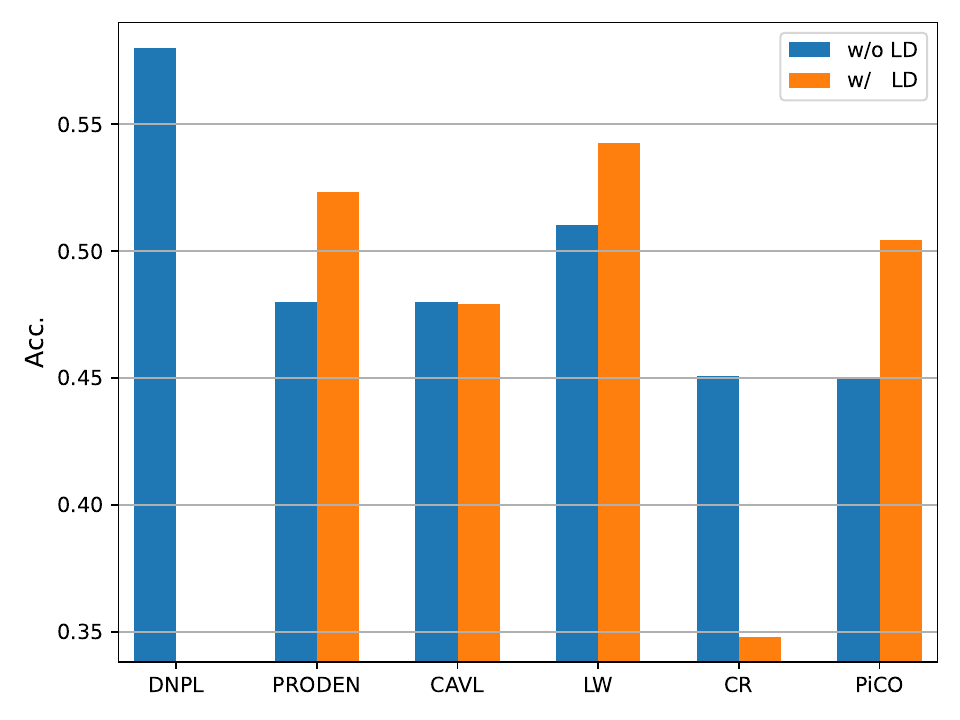} 
    \caption{SEED-IV: Average performance of the PLL methods across $6$ ambiguously labeled scenarios with classical PLL setup ($q\in \{0.2, 0.4, 0.6, 0.8, 0.9, 0.95 \}$).}
    \label{comparison (seed-iv)}
    \end{center}
\end{figure}
\begin{figure}[htp!]\color{black}
    \begin{center}
    \includegraphics[width=0.8\columnwidth]{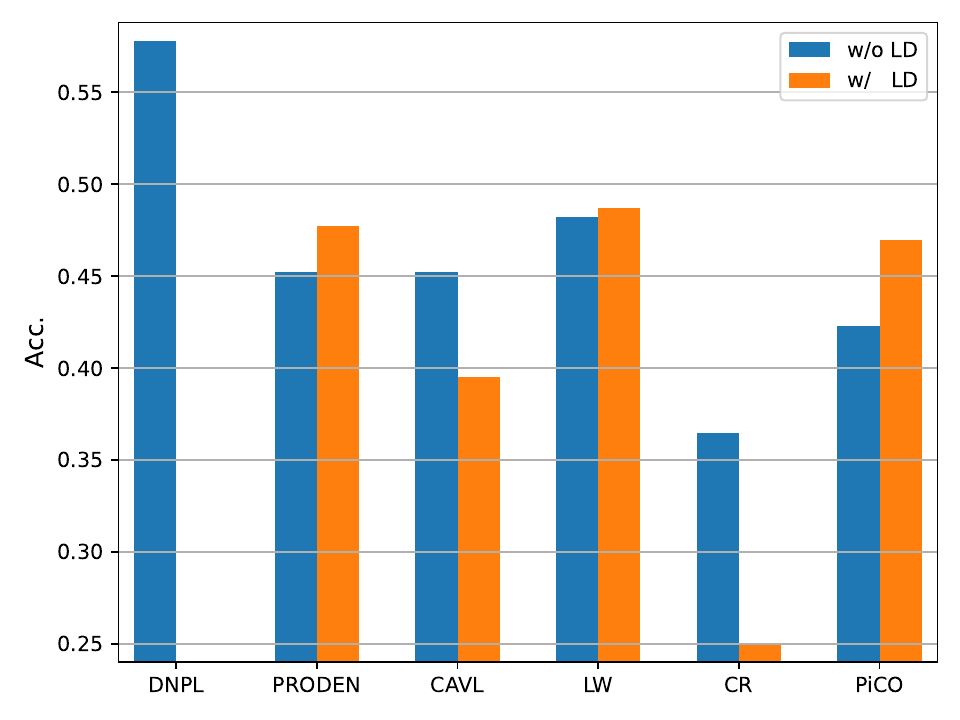} 
    \caption{SEED-V: Average performance of the PLL methods across $6$ ambiguously labeled scenarios with classical PLL setup ($q\in \{0.2, 0.4, 0.6, 0.8, 0.9, 0.95 \}$).}
    \label{comparison (seed-v)}
    \end{center}
\end{figure}

Furthermore, we compare the average performance across $q\in\{0.2, 0.4, 0.6, 0.8, 0.9, 0.95 \}$ for all the PLL methods. \textcolor{black}{As illustrated in Figures \ref{comparison (seed-iv)} and \ref{comparison (seed-v)}, in both the datasets, DNPL shows the best performance, followed by LW, when label disambiguation is not used. With label disambiguation, LW outperforms the rest, followed by PRODEN. We also observe that label disambiguation methods do not improve the model performance in CAVL \cite{zhang2022exploiting} and CR \cite{wu2022revisiting}. In SEED-IV, label disambiguation enhances performance for the LW method \cite{wen2021leveraged}, whereas, in SEED-V, it has a negligible impact on the average performance.}

\begin{figure}\color{black}
    \begin{center}
    \includegraphics[width=0.8\columnwidth]{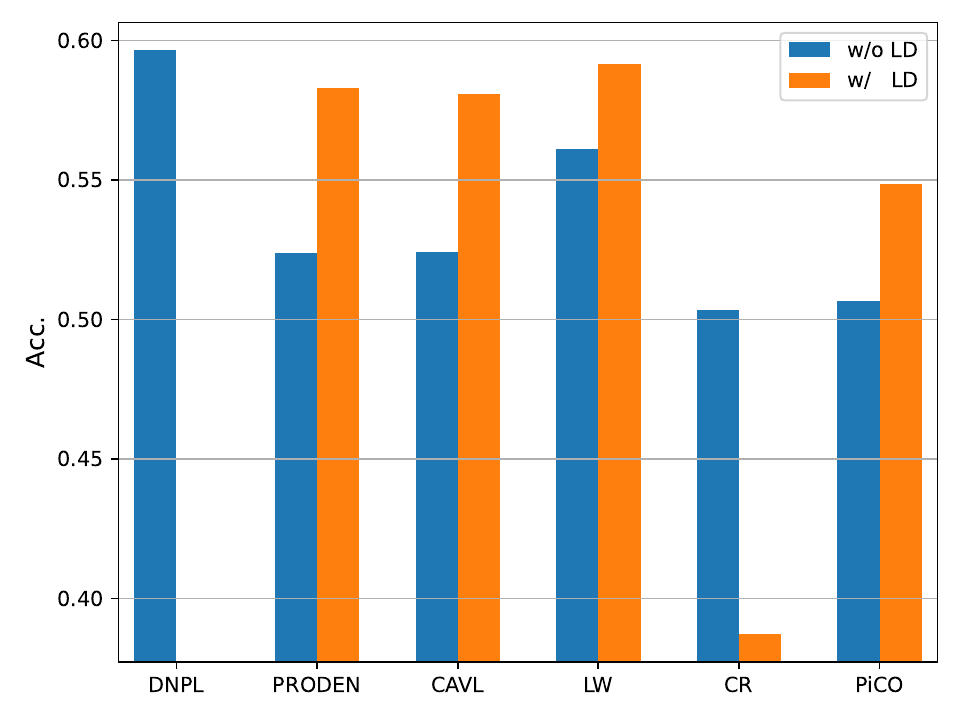} 
    \caption{SEED-IV: Performance of the PLL methods in the experiments with circumplex PLL setup.}
    \label{barplot/circumplex (seed-iv)}
    \end{center}
\end{figure}
\begin{figure}
    \begin{center}\color{black}
    \includegraphics[width=0.8\columnwidth]{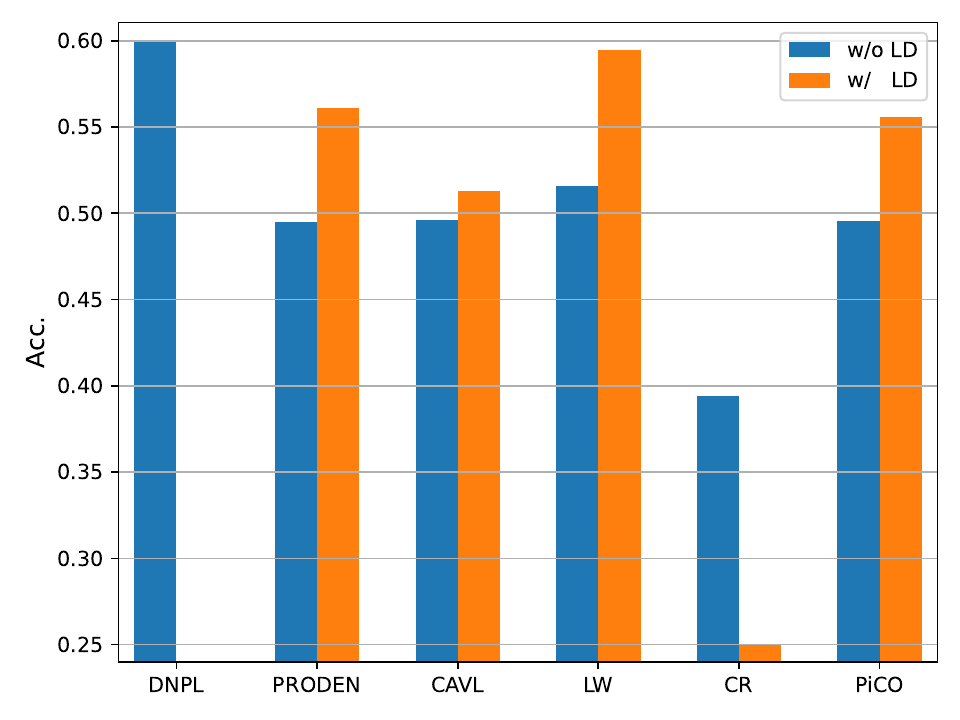} 
    \caption{SEED-V: Performance of the PLL methods in the experiments with circumplex PLL setup.}
    \label{barplot/circumplex (seed-v)}
    \end{center}
\end{figure}

In LW \cite{wen2021leveraged}, we further evaluate the method using both sigmoid and cross-entropy losses with three suggested values for the loss leveraging parameter ($\beta = 0, 1, 2$) (discussed in section \ref{method: LW}). This experiment is performed as these parameters are deemed important in the original LW paper \cite{wen2021leveraged}. \textcolor{black}{As shown in Tables \ref{table: LW (seed-iv)} and \ref{table: LW (seed-v)}, across both datasets, LW with cross-entropy loss consistently outperforms LW with sigmoid loss in all the $\beta$ and label disambiguation settings, across all the ambiguously labeled scenarios. LW with cross-entropy loss achieves the best results when more weights are assigned for learning non-candidate labels ($\beta = 2$).}

In PiCO \cite{wang2022pico}, we evaluate the impact of two major components, namely Contrastive Learning (CL) and label disambiguation (discussed in section \ref{method: PiCO}). \textcolor{black}{As shown in Table \ref{table: PiCO (seed-iv)}, in SEED-IV, PiCO performs the best using label disambiguation but without CL, across the majority of scenarios ($q\in\{0.2, 0.6, 0.8, 0.9, 0.95\}$) and ranks the second best when $q=0.2$.
As indicated in Table \ref{table: PiCO (seed-v)}, in SEED-V, we find that CL and label disambiguation are beneficial in cases where the candidate labels are less ambiguous ($q\in\{0.2, 0.4, 0.6\}$).} PiCO performs the best without CL when label ambiguity increases ($q\in\{0.8, 0.9\}$), and obtains the top performance without both components when label ambiguity is very high ($q=0.95$).

\subsection{Performance in the Circumplex PLL Setup}
We evaluate the performance of the existing PLL methods with the candidate labels generated based on the similarity scores between two emotions. \textcolor{black}{As demonstrated in Figures \ref{barplot/circumplex (seed-iv)} and \ref{barplot/circumplex (seed-v)}, in both the datasets, in contrast to the role of label disambiguation in PLL methods under classical setup (uniform distribution of candidate labels), label disambiguation improves model performance for the majority of methods under circumplex-based setup.} In particular, the LW method with label disambiguation closely approaches the best result. The only exception is in CR method \cite{wu2022revisiting}, where label disambiguation has a negative impact on model performance. We believe that compared to the classical experiments, label disambiguation is less likely to be misled by false positive labels and allows the model to focus on the candidate labels which are closer to the ground truth in circumplex-based experiments. Therefore, the label disambiguation process is helpful in identifying the ground truth from candidate labels, thus improving model performance. \textcolor{black}{Furthermore, we discover that DNPL \cite{seo2021power} and LW \cite{wen2021leveraged} with the label disambiguation process are able to approach the performance of the \textit{fully supervised learning} method ($62.11\%$ in SEED-IV and $63.08\%$ in SEED-V), addressing the challenge of ambiguous EEG labels in  circumplex-based affective computing.} We also observe that DNPL \cite{seo2021power} performs the best in both classical and circumplex-based experiments, demonstrating its effectiveness in handling candidate labels with various ambiguity levels.

\begin{figure*}
    \begin{center}
    \includegraphics[width=2.0\columnwidth]{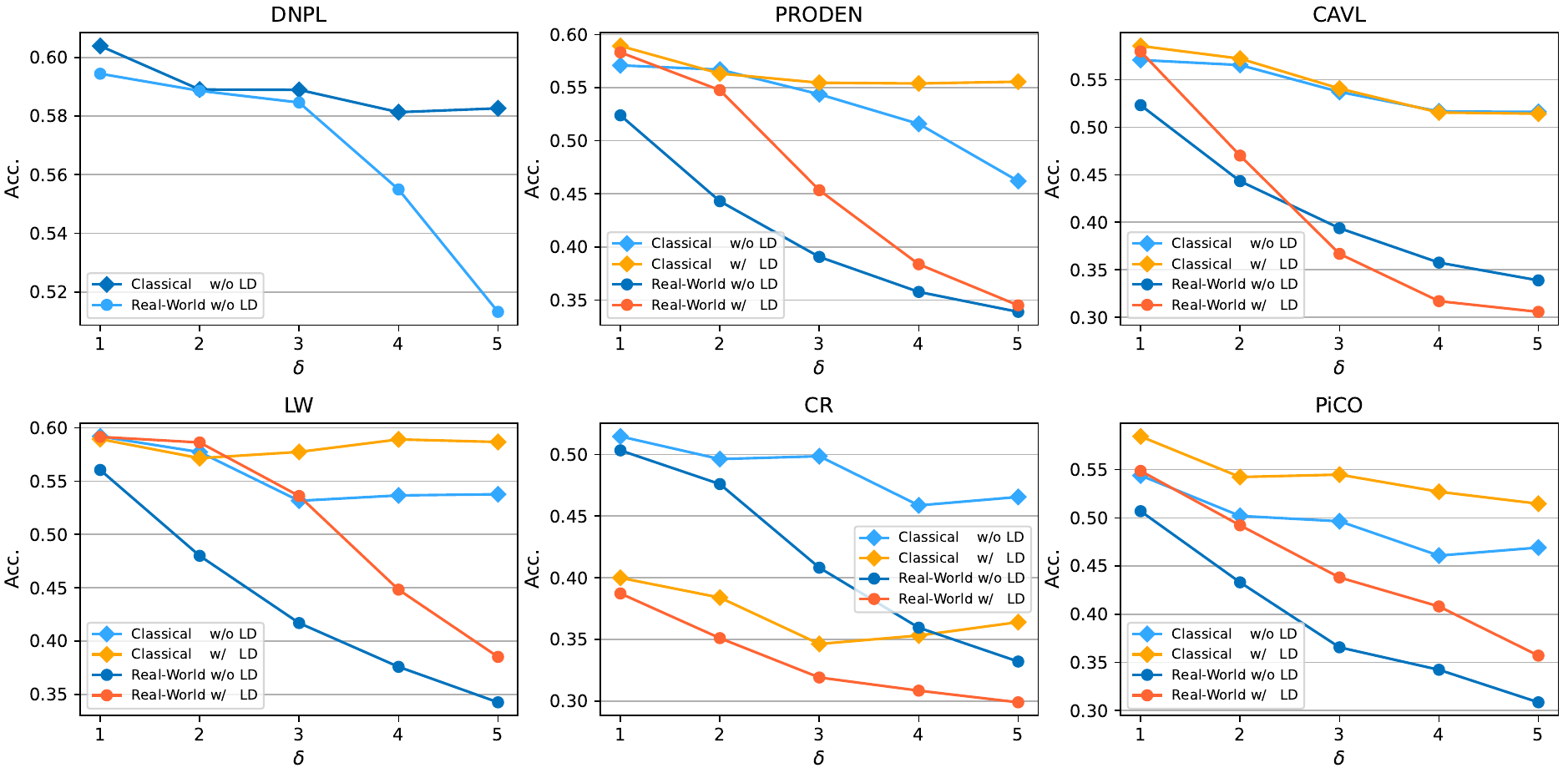} \color{black}
    \caption{SEED-IV: Performance comparison of all PLL methods in both classical and real-world setups. In the real-world PLL setup, we applied $\delta$ scaling on distances between emotions with $\delta \in \{1, 2, 3, 4, 5\}$ to gradually increase similarity scores, therefore increasing the averaged label ambiguity levels $q$. For a fair comparison, we applied the same ambiguity level with $q \in \{0.35, 0.54, 0.65, 0.72, 0.76\}$ in the classical PLL setup.}
    \label{delta (seed-iv)}
    \end{center}
\end{figure*}
\begin{figure*}
    \begin{center}
    \includegraphics[width=2.0\columnwidth]{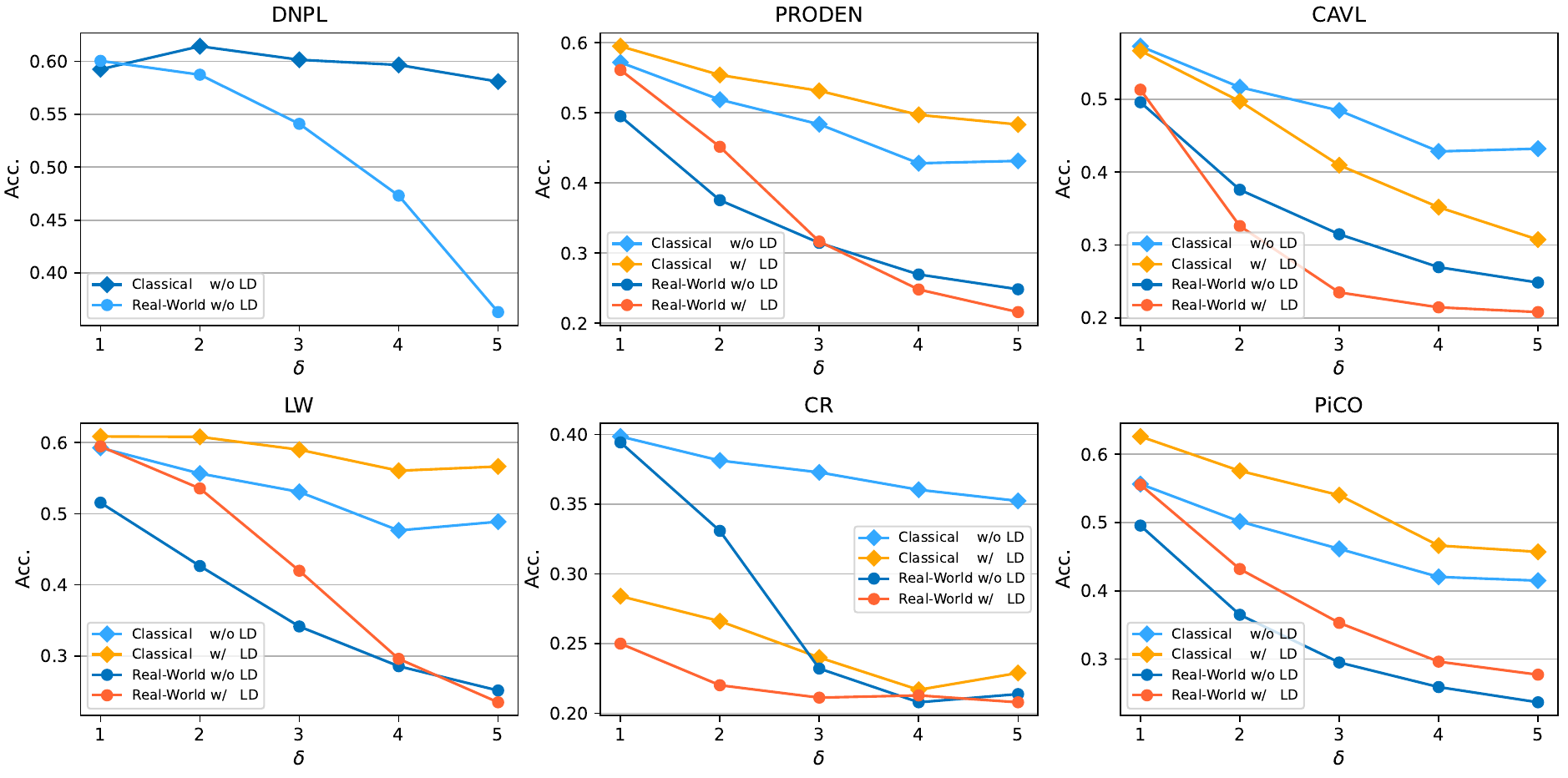} \color{black}
    \caption{SEED-V: Performance comparison of all PLL methods in both classical and real-world setups. In the real-world PLL setup, we applied $\delta$ scaling on distances between emotions with $\delta \in \{1, 2, 3, 4, 5\}$ to gradually increase similarity scores, therefore increasing the averaged label ambiguity levels $q$. For a fair comparison, we applied the same ambiguity levels with $q \in \{0.40, 0.59, 0.69, 0.75, 0.78\}$ in the classical PLL setup.}
    \label{delta (seed-v)}
    \end{center}
\end{figure*}

\subsection{\textcolor{black}{Performance in the Real-World PLL Setup}} \label{performance: real_world_setup}
\textcolor{black}{We evaluate the performance of all PLL methods in the real-world setup with $\delta \in \{1, 2, 3, 4, 5\}$. 
Furthermore, to enable a fair comparison between classical and real-world setups, as described in Section \ref{real_world_setup}, we apply $q \in \{0.35, 0.54, 0.65, 0.72, 0.76\}$ for the SEED-IV dataset \cite{zheng2018emotionmeter}, and $q \in \{0.40, 0.59, 0.69, 0.75, 0.78\}$ for SEED-V \cite{liu2021comparing}, respectively. As illustrated in Figures \ref{delta (seed-iv)} and \ref{delta (seed-v)}, PLL methods generally perform better in the classical PLL setup than in the real-world PLL setup, particularly at higher ambiguity levels. For example, the performance of DNPL \cite{seo2021power} drops sharply when $\delta>3$ in the real-world PLL setup, while it remains consistently high in the classical PLL setup. This is likely due to the exceptionally high similarity scores among closely related emotions in the real-world setup at higher ambiguity levels ($q$). 
Specifically, when $\delta>3$, the real-world setup presents an increased challenge due to significantly greater ambiguity among similar emotions compared to the overall average ambiguity. For instance, as shown in Figure \ref{emotion_pair_delta}, at $\delta=4$, ambiguity levels ($\gamma_{\delta}$) exceed $0.95$ for the `fear'-`disgust' and `sad'-`disgust' pairs, while the average ambiguity levels ($q$) are $0.72$ and $0.75$ in SEED and SEED-IV, respectively. These extremely high ambiguity levels among closely related emotions pose a substantial challenge for PLL methods. In contrast, the classical PLL setup is less challenging due to both the lower ambiguity level ($q$) and the uniform distribution of ambiguity.}

\textcolor{black}{In the real-world PLL setup, we observe that as $\delta$ increases, the positive effects of the label disambiguation process weaken in some cases. For example, in both the LW \cite{wen2021leveraged} and PRODEN \cite{lv2020progressive} methods, the benefits of label disambiguation diminish when $\delta\geq3$. This is primarily due to the increased difficulty of distinguishing closely related emotions when their similarity scores are extremely high (e.g., $\gamma_{\delta}>0.9$), which can even lead to negative effects at higher $\delta$ values, as seen in the performance of the PRODEN \cite{lv2020progressive}, CAVL \cite{zhang2022exploiting} and LW \cite{wen2021leveraged} on the SEED-V dataset. Interestingly, PiCO \cite{wang2022pico} maintains the positive effect of label disambiguation as $\delta$ increases, possibly due to the impact of contrastive learning.}

\textcolor{black}{Furthermore, we assess the impact of the label disambiguation process across datasets. Compared to SEED-IV, SEED-V includes an additional emotion class, `disgust', yielding two extra emotion pairs (`fear'-`disgust' and `sad'-`disgust') with the highest $\gamma_{\delta}$ values among the $10$ emotion pairs, particularly when $\delta\geq3$ (Figure \ref{emotion_pair_delta}). This increased challenge is evident in the performance of the PRODEN, CAVL, and LW methods, as shown in Figures \ref{delta (seed-iv)} and \ref{delta (seed-v)}. Specifically, in SEED-IV, PRODEN with label disambiguation consistently outperforms its counterpart without label disambiguation, whereas in SEED-V, PRODEN with label disambiguation performs worse when $\delta>3$. Similarly, LW with label disambiguation consistently performs better in SEED-IV, but in SEED-V, it performs worse when $\delta=5$. For CAVL, label disambiguation leads to worse performance in SEED-IV when $\delta>2$, while in SEED-V, this effect appears when $\delta>1$. 
Overall, DNPL \cite{seo2021power} consistently performs the best among PLL methods in both classical and real-world setups across datasets.
}

\subsection{\textcolor{black}{Impact of Ground Truth Reliability}} \label{reliability}
\textcolor{black}{The SEED-IV and SEED-V datasets used in this study benefit from rigorous data collection protocols including the careful selection of stimuli to evoke specific emotions and the use of facial expressions and physiological signals to reduce dependence on self-reported labels. However, emotion annotation remains inherently subjective, which can lead to inconsistencies in ground truth labels. To address this challenge, our study employs PLL to incorporate multiple candidate labels during training to reduce reliance on any single, potentially unreliable label, thus enhancing the robustness of the methods against noisy annotations. 
We acknowledge that obtaining perfect ground truth labels for emotion recognition would require additional sensing modalities, such as functional MRI \cite{morgenroth2023probing} or biochemical analysis (e.g., cortisol measurements) \cite{ma2017neural}, to further minimize subjectivity. However, these techniques are limited in practical applications due to their high cost, restricted accessibility, and invasive nature.}

\section{Conclusion} \label{sec:conclusion}
In this study, we tackle the challenge of ambiguous EEG labeling in emotion recognition tasks by adapting and implementing state-of-the-art \textit{partial label learning} methods. These methods were originally developed for computer vision applications. We conduct extensive experiments with six PLL frameworks across $6$ scenarios with different label ambiguities (across $q\in \{0.2, 0.4, 0.6, 0.8, 0.9, 0.95 \}$). \textcolor{black}{We also design circumplex-based and real-world scenarios where candidate labels are generated based on similarities among emotions instead of a uniform distribution assumption.} \textcolor{black}{We evaluate the performance of all the PLL methods and investigate the importance of the label disambiguation process in classical, circumplex-based and real-world experiments on two large publicly available datasets, SEED-IV and SEED-V with $4$ and $5$ emotion categories, respectively. The results show that, in the majority of cases, the label disambiguation process improves the model performance in circumplex-based experiments while causing performance degradation in the classical experiments. We believe the false positive labels generated with a uniform distribution would mislead the label disambiguation process, while the label disambiguation process is provided with better guidance when the candidate labels are generated based on the similarities with the ground truth. These results indicate the potential of using label disambiguation-based PLL frameworks for circumplex-based emotion recognition experiments.} 
\textcolor{black}{Furthermore, we revisited Russell's circumplex model and introduced $\delta$ scaling to control label ambiguity by adjusting the similarity scores for emotion pairs (e.g., `happy'-`fear'), enabling a PLL performance comparison between the classical and real-world PLL setups. The results shows that, when $\delta\leq5$, corresponding to an average ambiguity level of $q<0.8$, PLL methods generally perform better in a classical PLL setup, which is less challenging due to its uniform distribution of ambiguity. In contrast, the real-world setup poses a more realistic challenge, characterized by high ambiguity levels among closely related emotions. Our analysis indicates that $\delta>3$ leads to impractical scenarios where emotions become overly similar, compromising the model's ability to reflect real-world emotional relationships described by Russell's circumplex model. This threshold highlights the limitations of PLL methods and the reduced effectiveness of the label disambiguation process when similarity scores between emotions become excessively high, as demonstrated by the degraded performance of these methods. Therefore, we chose not to explore scenarios with $\delta>5$, as they would not reflect realistic relationships among emotions.}
\textcolor{black}{Overall, DNPL achieves the best performance and approaches \textit{fully supervised} learning in all the classical, circumplex-based and real-world experiments, addressing the challenge of ambiguous EEG labeling in emotion classification tasks.}

\noindent \textbf{Limitations and Future Work:}
\textcolor{black}{Our work targets scenarios where emotion labels are categorized (e.g., `happy', `sad', `neutral'), which corresponds to discrete models of affect. However, continuous models of affect, which represent emotions along continuous dimensions (as used in datasets such as DEAP \cite{koelstra2011deap}, MAHNOB-HCI \cite{soleymani2011multimodal}, DREAMER \cite{katsigiannis2017dreamer}, and AMIGOS \cite{correa2018amigos}) are not suitable for our current evaluations. Adapting PLL to continuous models would require substantial modifications, particularly in the label disambiguation process, which would need to predict continuous outputs instead of discrete labels. For future work, this adaptation can be explored by modifying the loss functions and disambiguation processes to handle continuous values rather than categorical labels. More advanced architectures can also be explored to investigate the impact of different architectural choices in this context.}

\bibliographystyle{IEEEbib}
\bibliography{refs}

\end{document}